\documentclass[twocolumn]{article}

\usepackage{arxiv}

\usepackage[utf8]{inputenc} 
\usepackage[T1]{fontenc}    
\usepackage{hyperref}       
\usepackage{url}            
\usepackage{booktabs}       
\usepackage{amsfonts}       
\usepackage{nicefrac}       
\usepackage{microtype}      
\usepackage{graphicx}
\usepackage{doi}
\usepackage{textcomp}
\usepackage{multirow}
\usepackage[abs]{overpic}
\usepackage[dvipsnames]{xcolor}
\usepackage{float}
\usepackage{tikz}
\usepackage{amsmath}
\usepackage{pdfpages}
\usepackage{listings}
\usepackage{comment}
\usepackage{ftnright}
\usepackage{enumitem}
\usepackage{adjustbox}
\usepackage{subcaption}
\usepackage{pgfplots}

\usepackage[
backend=bibtex,
style=numeric,
citestyle=numeric
]{biblatex}
\setitemize{noitemsep,topsep=0pt,parsep=0pt,partopsep=0pt,leftmargin=*}
\setlist[enumerate]{noitemsep,topsep=0pt,parsep=0pt,partopsep=0pt,leftmargin=*}

\newcommand{\B}[1]{\textbf{#1}}

\DeclareMathOperator*{\argmax}{argmax}
\def\ours{\makebox[0pt]{\hspace{-4em}Ours}}

\title{MDMMT: Multidomain Multimodal Transformer for Video Retrieval}

\author{
	Maksim Dzabraev$^{1,2}$, Maksim Kalashnikov$^{1}$, Stepan Komkov$^{1,2}$, Aleksandr Petiushko$^{1,2}$ \\
	$^1$Lomonosov Moscow State University \\
	$^2$Huawei Moscow Research Center \\
	dzabraev.maksim@intsys.msu.ru, kalashnikov.maxim@intsys.msu.ru, \\
	stepan.komkov@intsys.msu.ru, petyushko.alexander1@huawei.com
}

\hypersetup{
pdftitle={Multidomain multimodal video retrieval},
pdfauthor={M.Dzabraev, M.Kalashnikov, S.Komkov, A.Petiushko},
pdfkeywords={video, language, retrieval, multi-modal, multimodality, cross-modal, temporality, transformer, attention, text to video retrieval},
}

\begin{document}
\twocolumn

\twocolumn[
  \begin{@twocolumnfalse}
    \maketitle
  \end{@twocolumnfalse}
]

	\centerline{\bf ABSTRACT}
We present a new state-of-the-art on the text to video retrieval task on MSRVTT and LSMDC benchmarks where our model outperforms all previous solutions by a large margin.
Moreover, state-of-the-art results are achieved with a single model on two datasets without finetuning. This multidomain generalisation is achieved
by a proper combination of different video caption datasets. We show that training on different datasets can improve test results of each other. Additionally we check
intersection between many popular datasets and found that MSRVTT has a significant overlap between the test and the train parts, and the same situation is observed for ActivityNet.


\section{Introduction} \label{sec:introduction}
Video is a quite popular data format, 500+ hours of video are uploaded on YouTube every minute. Many personal mobile phones
have gigabytes of video. Since video format gets more popular every year the importance of modern search methods is increasing as well.

In this work we present our research about text to video retrieval task. In this task system should return for a given textual query the most relevant video
segments from a gallery. The query is a textual description of what we want to find in the video. The query may describe objects, actions, sounds, ..., and
relations between them.

Such search methods are a promising direction for mobile devices because every year manufacturers increase the available memory on devices.
The large part of the memory is filled by media data. For end users it is getting difficult to search for a video made one or two years ago.
But users can easily describe the content of the video using natural language, which can be effectively used as a search query.

\begin{figure}[H]
  \begin{tabular}{c@{\hspace{0.5em}}c}
    \begin{minipage}{0.48\columnwidth}%
      \includegraphics[width=0.95\textwidth,bb=0 0 306 207]{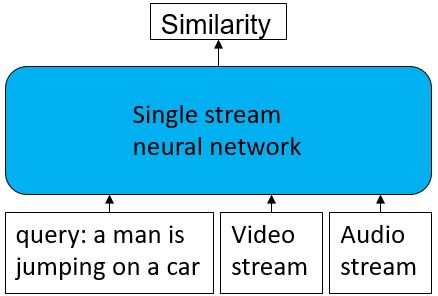}
    \end{minipage}
    &
    \begin{minipage}{0.48\columnwidth}%
      \includegraphics[width=0.95\textwidth, bb=0 0 347 252]{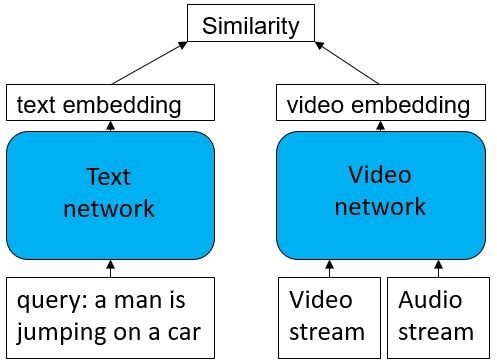}
    \end{minipage} \\\\
    \begin{minipage}{0.48\columnwidth}
      \begin{subfigure}{\columnwidth}
	\caption{Scheme for a single-stream neural network.}\label{fig:sstream_network}
      \end{subfigure}
    \end{minipage} & 
    \begin{minipage}{0.48\columnwidth}
      \begin{subfigure}{\columnwidth}
	\caption{Scheme for a two-stream neural network.}\label{fig:tstream_network}
      \end{subfigure}
    \end{minipage} \\
  \end{tabular}

\caption{Two types of fusion} \label{fig:1}
\end{figure}

There are two major directions which allow calculate the relevance between a textual search query and a video segment. The first direction is single stream
approaches \cite{sun2019videobert}, where a query and a video together are given to a network and then become fused from the beginning of the processing. The schematic illustration of this approach
is presented in Fig.~\ref{fig:sstream_network}.

This type of approaches have access to all input data from the beginning of its processing
and can make a strong verdict about data. But these approaches have a significant drawback because it is not scalable:
every new query the search system should calculate the full forward pass for this query and for each video segment from the gallery.

Another direction is two stream neural networks \cite{mithun2018learning}, \cite{gabeur2020multimodal}, where a textual query and a video are processed by two different neural networks. As
a result the networks produce embeddings inside the same embedding space, where semantically close textual queries and video segments will
be placed next to each other. The schematic illustration is presented in Fig.~\ref{fig:tstream_network}.

The two stream approach is scalable, it allows to precompute video embeddings for all videos from the gallery, and to do only one forward pass with the text network for each new
query and then to compute the cosine similarity between the new query embedding and all precomputed embeddings.

To make a strong video retrieval solution it is important to show to the model a lot of situations, actions and objects from real life.
There exist a lot of video datasets, but none of them cover a significant portion of real life situations. One of the first steps to
tackle this problem is to formulate the rules for combining different existing datasets to a single large train database.

Text to video retrieval is a modern direction, where one of the first works was published at 2016 \cite{torabi2016learning}. One of the most universal solution for video retrieval task
is Multi Modal Transformer~\cite{gabeur2020multimodal} architecture which uses BERT~\cite{devlin2019bert} backbone for a video network. It allows in a natural way to process the temporal dependencies inside the multi modal
data source.

To train a text to video retrieval neural network the training database should consists of pairs: (a video segment, a textual description of this video segment). Traditionally such sort of datasets was created for a video captioning task.
But it turns out that these datasets perfectly can be used for a video retrieval task.
One of the first video captioning dataset was MSVD, which was created in 2010. Today there exist more than a dozen of different video captioning datasets.

The most popular datasets for text to video retrieval is MSRVTT~\cite{xu2016msr-vtt}, ActivityNet~\cite{krishna2017dense} and LSMDC~\cite{rohrbach2016movie}.
Many researchers test their solutions mostly on these three datasets.

Our main contributions in this work are the following:
\begin{itemize}
  \item We present a new state-of-the-art (SotA) result on MSRVTT and LSMDC benchmarks;
  \item We present a model which shows good results on three different benchmarks without finetuning: MSRVTT (SotA), LSMDC (SotA) and ActivityNet at the same time;
  \item We present a practical approach which helps us to find the overlap between the train and the test parts of used datasets.
\end{itemize}

\section{Related work}
\subsection{Datasets}
\label{sect:datasets}
MSRVTT~\cite{xu2016msr-vtt} was created in 2016. This dataset is traditionally used by researchers as the main dataset for testing text to video retrieval models. This dataset consists of 10k video segments, each segment has 20 captions.
The authors collected 257 popular search queries and gathered from YouTube 118 most relevant videos for each of them.
The dataset has 42 hours of video. The captions were made by 1327 amazon workers.

Today there are three different test/train splits. The official split is called \textbf{full} split, where the train part has 7k videos
and the test part has 3k videos. There are two important properties of this split: 1. there are no two video segments cropped from the same video so as
the first segment is placed in the train part and the second segment is placed in the test part; 2. there are no two video segments,
retrieved from the same query so as the first one is placed in the train part and the second one is placed in the test part.

Another two splits are called \textbf{1k-A}~\cite{yu2018joint} (sometimes called jsfusion) and \textbf{1k-B}~\cite{miech2020learning} (sometimes called miech).
Both of them have different 1k videos for testing.
They were created by randomly sampling 1k videos from the original test part (full split). 1k-A train part consists of the original train split and the
rest of the videos from the test part, so it has 1k videos for the test part and 9k videos for the train part. 1k-B has 1k videos for the test part and 6.5k videos for the train.
Additionally both splits use only one caption per segment (instead of 20 captions).

Unfortunately 1k-A and 1k-B mixed up the train and test parts. This led to violation in properties 1. and 2. which the full split satisfies.

Another problem is that all these splits have the overlap between the test and train parts, see~\ref{sect:cleaning_res} for details. 
To be strict we remove the overlap between the test part and the train part of MSRVTT full split.
We called this split MSRVTT \textbf{full clean}, and refer to it as M$_c$.
It is worth to mention that we do not modify the test part, we only remove some videos from the train part.

%

%
%
%
%

The Large Scale Movie Description Challenge (LSMDC)~\cite{rohrbach2016movie} is the extension of two independent datasets:
MPII Movie Description Dataset (MPII-MD)~\cite{rohrbach2015dataset}, and Montreal Video Annotation Dataset (M-VAD)~\cite{torabi2015using}.

Video segments for this dataset were cropped from movies, where movie textualized transcriptions were used as captions. A movie transcription
is an audio description of a video segment that helps blind people to watch movies by describing what happens, who appears in this time,
what is on background right now and so on.

In this work for testing we use LSMDC public test, which consists of 1k video segments.

%
%
%
%
%

ActivityNet captions dataset \cite{krishna2017dense} consists of 20k videos and 100k captions, where captions cover the full video length for the most of videos, and neighbour captions may intersect. 
The annotations were made with Amazon Mechanical Turk.

The situation when some video segments may overlap makes a problem for text to video retrieval testing.
Suppose we have two video-caption pairs $(S_1, C_1)$ and $(S_2, C_2)$ where the video segment $S_1$ has a non empty overlap with the video segment $S_2$.
Now suppose that for query $C_1$ the system returns the video segment $S_2$. Is it mistake or not? What to do in this case?

Many previous works used ActivityNet test dataset in a paragraph retrieval mode. In this mode all captions for all video segments are concatenated,
then the concatenated text is used as a textual query and the whole video should be retrieved for this query.
Such mode has two drawbacks. The first one is that paragraph retrieval is not a classical video retrieval mode, it is another task. One can ask:
if a model is good in paragraph retrieval will it be good for video retrieval? The second drawback is that queries will be long,
video segments will be long (compared to a classical video retrieval mode). This issue requires to enlarge the input for the model.

Another way to use the test part of ActivityNet is just to sample once a single random segment from each video. As a result we will have non intersected video segments
and captions with usual length. We use ActivityNet test part in this way. We take all videos from val1 and val2 parts, and sample a single random segment from each video.
All results on ActivityNet are reported on this split.

Additionally in this work the following datasets are used:
NIST TRECVID Twitter vines~\cite{2020trecvidawad},
TGIF~\cite{tgif-cvpr2016},
MSVD~\cite{chen-dolan-2011-collecting},
YouCook2~\cite{zhou2018weaklysupervised},
Something-something V2~\cite{goyal2017something},
Kinetics 700~\cite{smaira2020short},
HowTo100M~\cite{miech19howto100m}.

\subsection{Prior Art} \label{ssec:mmt}
A dominant approach to train video retrieval models is contrastive learning. The idea of this approach is that we have a set of pairs $(\text{video}_i, \text{text}_i)$ and
elements of each pair should be placed next to each other in some metric space: $\text{distance}(\text{video}_i, \text{text}_i) \rightarrow 0$, at the same time the element $\text{video}_i$
should be far from all other $\text{text}_j), j\neq i$: $\text{distance}(\text{video}_i, \text{text}_j) \rightarrow +\infty$. The bi-directional max-margin ranking loss~\cite{Karpathy2014DeepFE} represents this idea.

When training data have a lot of noise the MIL NCE loss~\cite{miech2020endtoend} can be applied in the training procedure. Suppose that we know that a video$_i$ should be close to one of (or several) texts text$_{i1}$, ..., text$_{ik}$.
This approach tries to reduce the distance between the video$_i$ and all text$_{i1}$, ..., text$_{ik}$ at the same time.

All video captions datasets have the following problem. Suppose the distance between $(\text{video}_i, \text{text}_i)$ is to be minimized while the distance between $(\text{video}_i, \text{text}_j), j\neq i$ is to be maximized, but
$\text{text}_i$ and $\text{text}_j$ are quite similar (from the semantical point of view). Maybe the optimal scenario in this situation is to minimize the distance between $(\text{video}_i, \text{text}_j), j\neq i$.
In~\cite{patrick2021supportset} the authors show the approach which deals with this problem.

As far as an input video is the temporal sequence of tokens (frames or video segments) it is important to efficiently aggregate the information from all tokens. Many ideas for such aggregation in the previous works are borrowed 
from the natural language processing. Convolution filters for aggregation are used in~\cite{patrick2021supportset}, a transformer encoder as a video aggregator is used in~\cite{gabeur2020multimodal},
many different aggregation functions are tested in~\cite{portilloquintero2021straightforward}.

We think that the most promising aggregation method is Multi Modal Transformer (MMT)~\cite{gabeur2020multimodal}.
MMT is a two stream solution designed for a text to video retrieval task.
The extraction of features from the input video stream is done in the following way.
An input video is preprocessed by several pretrained frozen
neural networks (these networks are called experts). Original solution uses seven modalities: motion, RGB, scene, face, OCR, speech, audio, and one pretrained network for each modality is used.
The motion modality is processed with video recognition networks like S3D, SlowFast, irCSN, where several input frames are used as a single input.
The RGB modality uses a single frame as an input. The audio modality uses the raw input sound from a video.
After embeddings are extracted from input data by these experts, it will be augmented by adding positional encoding tokens (representing time) and
expert tokens.
Then the augmented embeddings are passed through MMT backbone. MMT backbone is a standard transformer encoder architecture.
Each input modality produces one embedding, so in total there are seven output embedding from MMT.

For encoding the textual query the authors use pretrained BERT model where the output [CLS] token is used. The output is postprocessed with shallow networks (one network per modality) to extract
the modality related information, in total seven feature vectors will be produced.
In addition to embeddings from the text query seven weights representing how much the query describes one of seven modalities are produced. For example, if a query does not represent the sound,
the small weight for the audio modality should be produced.

The final similarity score is done by a sum of seven weighted dot products of embeddings.

The MMT is trained with the bi-directional max-margin ranking loss~\cite{Karpathy2014DeepFE}:
\[
	\frac{1}{B}\sum_{i=1}^{B} \sum_{j \neq i} \Big[ \max(0, s_{ij} - s_{ii} + m) + \max(0, s_{ji} - s_{ii} + m)\Big]
\]
where $B, s_{ij}, m$ represent the batch size, the similarity between the $i$-th query and the $j$-th video inside this batch, and some predefined margin correspondingly.

\section{Methodology}
Our work is mostly based on MMT. We use the same loss and a similar architecture, but with different hyperparameters.
In this work we study the following questions:

\begin{itemize}
	\item Which publicly available pretrained motion expert is the best for text to video retrieval nowadays, Sec.~\ref{sec:motion_experts}.
	\item How to combine several video caption datasets in order to train a strong model without specialisation for a particular dataset, Sec.~\ref{sec:dset_compose}.
	\item How to find and prevent the overlap between the test and train parts when combining datasets, Sec.~\ref{sec:tt_isect}.
\end{itemize}

\subsection{Motion experts} \label{sec:motion_experts}
The MMT video backbone does not process the raw input video stream, and instead the input video stream
is processed by one or more pretrained experts, where each expert produces time series of features.
The most important modality is motion: a motion expert processes several video frames as a single input unit and extracts the information
about actions and objects within a segment.

We may say that the motion modality is the basis of the MMT. If a motion expert doesn't extract some information, there is a high probability that
MMT won't know about some events in the video stream. That's why improving the motion expert is very important.

We consider several best solutions from Kinetics~\cite{kay2017kinetics} benchmark as well as several promising video recognition models
and check which one works in the best way as a motion expert.
We present all details in Sec.~\ref{ssec:video_experts}.

\subsection{Dataset creation} \label{sec:dset_compose}
	It is possible to train a video retrieval model by two means. The first way is the way of specialization for a single domain.
For example: create model that will work good only for MSRVTT benchmark (or domain) but at the same time this model will show poor results on other datasets (domains).
In this way MMT \cite{gabeur2020multimodal} was trained. The authors trained three different models for MSRVTT, ActivityNet and LSMDC datasets. Each of these
three networks works good on domain $X$ if and only if it was trained on $X$, but at the same time works poor on another domain $Y\neq X$.
A proof of this statement we provide in Tab.~\ref{tab:cross-dataset-test}.

The second way is to create a model that will work good for all domains at the same time. We use this way.

Obviously the model trained in the first way can't work good with real users, because the event when a user writes a
search query similar to some caption from a small train database is very rare.

The second drawback here is that each video retrieval
train dataset is not that big, and it causes the situation that model doesn't see many words and real life situations during training. For example, MSRVTT has
only 9k videos and 200k captions in total for training, obviously this is not enough to train a neural network that will know most of real life
situations, different items and persons. To tackle with this problem we can take several datasets with videos and captions and concatenate it.

\begin{figure}[h]
      \centering
\scalebox{0.90}{
\begin{tikzpicture}

\draw[->,ultra thick] (-0.2,0)--(6,0);
\draw[->,ultra thick] (0,-0.2)--(0,6);
\node[] at (3,-0.6) {number of unique captions};
\node[] at (-0.5, 3) {\rotatebox{90}{total video duration}};

\filldraw[color=red] (5.0,0.4468689629600283) circle (0.38461538461538464);
\node[above, color=red] at (5.0,0.831484347575413) {MSRVTT};
\filldraw[color=blue] (2.0596715219085295,5.0) circle (0.38461538461538464);
\node[above, color=blue] at (2.0596715219085295,5.384615384615385) {ActivityNet};
\filldraw[color=cyan] (3.0762452796259665,1.2725879748861566) circle (0.11538461538461539);
\node[above, color=cyan] at (3.0762452796259665,1.387972590270772) {LSMDC};
\filldraw[color=magenta] (0.6935203500569442,0.11533771921202855) circle (0.07692307692307693);
\node[above, color=magenta] at (0.6935203500569442,0.1922607961351055) {Vines};
\filldraw[color=pink] (0.35059641551279747,0.7212969347454151) circle (0.11538461538461539);
\node[above, color=pink] at (0.35059641551279747,0.8366815501300305) {YC2};
\filldraw[color=violet] (3.766888449319667,1.1040190706769908) circle (0.5);
\node[above, color=violet] at (3.766888449319667,1.6040190706769908) {TGIF};
\filldraw[color=green] (1.9095186717017323,0.10215175014212363) circle (0.023076923076923078);
\node[above, color=green] at (1.9095186717017323,0.1252286732190467) {MSVD};
\node[] at (0.29970628783791886,-0.7em) {10K};
\node[] at (1.4985314391895943,-0.7em) {50K};
\node[] at (2.9970628783791886,-0.7em) {100K};
\node[] at (4.975124378109453,-0.7em) {166K};
\node[] at (-1.3em,5.0) {466h};
\node[] at (-1.3em,1.1040190706769908) {102h};
\node[] at (-1.3em,0.4468689629600283) {41h};
\end{tikzpicture}
}
  \caption{Radius of the ball represent the ``information size'' of dataset. The biggest balls have more diversity in data.}
  \label{fig:dataset_information_size}
\end{figure}
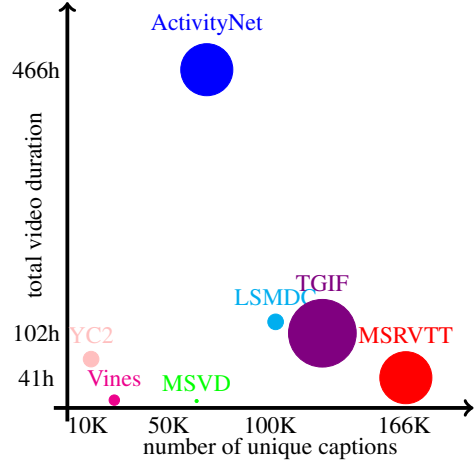

Different datasets have the different number of videos, the different number of captions per video, some datasets may have long captions, some may have short captions,
different rules for creating captions were used by human writers, and so on. Due to these factors some datasets may contain more information and require 
longer training time, some datasets may contain less information and require shorter training time. On the other hand, if we use long training time for
a small dataset it could lead to overfitting on this dataset (the data will be memorized). The "information sizes" of some used datasets are illustrated in Fig.~\ref{fig:dataset_information_size}.

Fig.~\ref{fig:dataset_information_size} is made with a simple algorithm. We take the original training procedure of MMT and for a given dataset we change the number of examples that
will be shown to a network during training. We define the radius of the ball as the number of training examples after which the performance gets saturated (i.e. increasing the training time
does not give the better model).

The key question is: what is the proper way for sampling examples from several datasets taking into account the different information size?

We use these obvious rules:

\begin{enumerate}
	\item If a dataset $X$ is larger than $Y$, we should sample from $X$ more often than from $Y$;
	\item Training on $X$ and $Y$ combined requires longer train than training solely on $X$ or $Y$;
	\item Training on $X$ and $Y$ combined may require a deeper model than for $X$ or $Y$.
\end{enumerate}

If we achieve the same results on $X$ after combining $X$ and $Y$ it is still good because model gets better on $Y$. 
Our experiments show that the proper usage of rules 1--3 often improves the results for a specific test dataset (e.g. MSRVTT) after extending the train dataset.

We managed to combine the following datasets: MSRVTT, ActivityNet, LSMDC, TwitterVines, YouCook2, MSVD, TGIF and Something to something V2 (SomethingV2).
In total we increase the number of video segments by 40 times and the number of unique captions by 4 times compared with MSRVTT dataset. In Tab.~\ref{tab:dataset_size} we summarize the sizes of used datasets.
We separate SomethingV2 dataset from all other datasets because: 1. all video segments are created artificially, 2. the structure of text captions is quite limited. At the same time videos for all other datasets
are collected from the Internet and captions being created by humans have quite a rich structure.

\begin{table}[h]
	\centering
	\begin{tabular}{ |l|cccc| }
		\toprule
		\multirow{3}{*}{Dataset}  & Num   & Num   & Num        & Has\\
								  & video & pairs & unique     & YouTube\\
								  &       &       & captions   & Id	    \\
		\midrule
		MSRVTT			& 10k		    & 200k			& 167k & Yes \\
		ActivityNet		& 14k		    & 70k			& 69k  & Yes \\
		LSMDC			& 101k		    & 101k			& 101k & No  \\
		TwitterVines    & 6.5k		    & 23k			& 23k  & No  \\
		YouCook2		& 1.5k		    & 12k			& 12k  & Yes \\
		MSVD			& 1.5k		    & 80k			& 64k  & Yes \\
		TGIF			& 102k          & 125k          & 125k & No  \\
		\textit{\bf Sum above}  & \textit{\bf 236k} & \textit{\bf 611k}         & \textit{\bf 561k}  & --- \\
		SomethingV2		& 193k		    & 193k			& 124k & No \\
		\textit{\bf Sum above}  & {\it\bf 429k}      & {\it\bf 804k}            & {\it\bf 685k} & --- \\
		\bottomrule
	\end{tabular}
	\caption{The "Num video" column represents the number of video clips in the dataset,
			 the "Num pairs" column represents the total number of video-caption pairs,
			 the "Num unique captions" column represents the number of unique captions in the dataset.}
	\label{tab:dataset_size}
\end{table}

\subsection{Intersection}
It is important to extend the training database carefully, not allowing the addition to the train part of video segments that already exist in the test part. 

To find the intersection between the test part and the train part we use the two stage filtration. The first stage is to use the YouTube ID, if it is available. We should
not allow to use in the test and train parts simultaneously any two video segments sampled from the same video. In the second stage we compute the similarity score between
each video from the test part and each video from the train part, then we manually assess the pairs with the highest scores.
In total we assessed more than 100K pairs of the most relevant segments, see Sec.~\ref{ssec:NDVS} for details.

We found the significant overlap between the MSRVTT 1k-A test and train parts, and the similar situation is with the 1k-B test and train parts and the less significant overlap is found between the MSRVTT full split test and train parts.
The similar situation is with the ActivityNet train and validation 1,2 parts.

Additionally we estimate (but did not find) the overlap between HowTo100M and MSRVTT, and found that it may be
significant. Our approach allows to approximately estimate the total number of videos in the intersection without finding the exact intersection, please see the details in Sec.~\ref{ssec:how_many_pairs}.
The similar estimation is for ActivityNet and Kinetics700, an our approximation shows that there may be a significant overlap, see all details in Sec.~\ref{ssec:NDVS}.

\section{Experiments}

\begin{table}[h]
	\centering
	\begin{tabular} {|l|l|}
		\toprule
		Abbreviate & Composition \\
		\midrule
		M				& MSRVTT full split\\
		M$_c$                           & MSRVTT full clean split, see Sec.~\ref{sect:datasets}\\
		M$_{\text{1k-A}}$		& MSRVTT 1k-A split\\
		M$_{\text{1k-B}}$		& MSRVTT 1k-B split\\
		A				& ActivityNet \\
		A$_{\text{val1}}$	        & ActivityNet val1 validation set\\
		A$_{\text{val2}}$	        & ActivityNet val2 validation set\\
		A$_{p/r}$			& ActivityNet paragraph retrieval, see Sec.~\ref{sect:datasets}\\
		L				& LSMDC \\
		K				& Kinetics700 \\
		V				& Twitter Vines \\
		Y				& YouCook2 \\
		HT100M				& HowTo100M \\
		\midrule
		\multirow{2}{*}{MALV}		& MSRVTT + ActivityNet + \\
						& LSMDC + TwitterVines \\
		\midrule
		\multirow{3}{*}{MALVYMT}	& MSRVTT + ActivityNet + \\
									& LSMDC + TwitterVines + \\
									& YouCook2 + MSVD + TGIF \\
		\midrule
		\multirow{4}{*}{MALVYMTS}	& MSRVTT + ActivityNet + \\
									& LSMDC + TwitterVines + \\
									& YouCook2 + MSVD + TGIF + \\
									& Something to Something V2 \\
		\bottomrule

	\end{tabular}
	\caption{The left column represents the abbreviate name for the set of datasets from the right column.}
	\label{tab:abbrev_names}
\end{table}

\subsection{Architecture}
We use exactly the same neural network architecture as original MMT \cite{gabeur2020multimodal}, our method is significantly based on their codebase. The difference is in the following:
1. we use the more aggressive dropout equals to 0.2 for the text BERT and the video BERT (against the original value of 0.1); 
2. we found that the deeper and wider transformer encoder for a video network gives better results~---
we use 6 layers and 8 heads for the motion only modality and 9 layers and 8 heads for the motion + audio setting (against 4 layer and 4 head in the original implementation).

\subsection{Stronger motion experts} \label{ssec:video_experts}
As the input data for MMT is embeddings from experts, the obvious question can arise: if a better expert is used, will we have a stronger model?
To answer this question we train MMT on MSRVTT dataset with the only motion modality. For motion experts we try several architectures
pretrained on different datasets, these models are presented in Tab.~\ref{tab:video_experts_fps32}. We take the architectures which show the best
results on Kinetics 400 benchmark having publicly available pretrained weights:
\cite{xie2018rethinking}
\cite{feichtenhofer2019slowfast}
\cite{tran2019video}
\cite{tran2018closer}
\cite{ghadiyaram2019largescale}
.

The results in Tab.~\ref{tab:video_experts_fps32} are made with the same hyperparameters as in \cite{gabeur2020multimodal}. For the train dataset we use only MSRVTT full clean split.
The first line in Tab.~\ref{tab:video_experts_fps32} represents the motion feature extractor from the original MMT paper.

\begin{table*}[h]
	\centering
	\begin{tabular}{|ll|l @{\hspace{1\tabcolsep}} l @{\hspace{1\tabcolsep}} l @{\hspace{1\tabcolsep}} l @{\hspace{1\tabcolsep}} l|}
		\toprule
			\multirow{2}{*}{Video expert}  & \multirow{2}{*}{Dataset} & \multicolumn{5}{c|}{Text $ \rightarrow$ Video} \\
				&  & R@1$\uparrow$	& R@5$\uparrow$	& R@10$\uparrow$	& MnR$\downarrow$	& MdR$\downarrow$ \\
		\midrule
		s3d                 & Kinetics 600		    & 7.7$_{\pm 0.1}$     & 24.0$_{\pm 0.2}$    & 34.9$_{\pm 0.2}$    & 129.6$_{\pm 1.0}$   & 23.7$_{\pm 0.5}$\\
		SlowFast 32x2 R101  & Kinetics 600		    & 9.3$_{\pm 0.1}$     & 27.5$_{\pm 0.1}$    & 39.1$_{\pm 0.1}$    & 110.8$_{\pm 1.1}$   & 18.7$_{\pm 0.5}$ \\ 
		ipCSN152            & IG65M			    & 9.5$_{\pm 0.1}$     & 27.9$_{\pm 0.2}$    & 39.6$_{\pm 0.2}$    & 106.1$_{\pm 1.1}$   & 18.0$_{\pm 0.0}$ \\
		ipCSN152            & IG65M $\rightarrow$ K400	    & 8.3$_{\pm 0.1}$     & 25.2$_{\pm 0.1}$    & 36.5$_{\pm 0.2}$    & 124.3$_{\pm 0.2}$   & 21.0$_{\pm 0.0}$ \\
		ipCSN152            & Sports1M			    & 7.4$_{\pm 0.2}$     & 22.4$_{\pm 0.1}$    & 32.7$_{\pm 0.2}$    & 140.6$_{\pm 1.0}$   & 27.0$_{\pm 0.0}$ \\
		ipCSN152            & Sports1M $\rightarrow$ K400   & 7.8$_{\pm 0.1}$     & 24.2$_{\pm 0.1}$    & 35.2$_{\pm 0.1}$    & 129.9$_{\pm 0.2}$   & 23.0$_{\pm 0.0}$ \\ 
		irCSN152            & IG65M			    & 9.5$_{\pm 0.1}$     & 27.9$_{\pm 0.2}$    & 39.5$_{\pm 0.2}$    & 105.5$_{\pm 0.4}$   & 18.0$_{\pm 0.0}$ \\
		irCSN152            & IG65M $\rightarrow$ K400      & 8.4$_{\pm 0.1}$     & 25.3$_{\pm 0.1}$    & 36.5$_{\pm 0.2}$    & 120.4$_{\pm 0.4}$   & 21.0$_{\pm 0.0}$ \\
		irCSN152            & Sports1M			    & 6.9$_{\pm 0.1}$     & 21.6$_{\pm 0.1}$    & 31.6$_{\pm 0.1}$    & 141.9$_{\pm 0.4}$   & 28.7$_{\pm 0.5}$ \\
		irCSN152            & Sports1M $\rightarrow$ K400   & 7.7$_{\pm 0.1}$     & 24.1$_{\pm 0.1}$    & 35.1$_{\pm 0.1}$    & 127.6$_{\pm 0.6}$   & 23.0$_{\pm 0.0}$ \\
		r(2+1)d 152         & IG65M			    & 5.7$_{\pm 0.1}$     & 18.5$_{\pm 0.1}$    & 27.8$_{\pm 0.1}$    & 178.5$_{\pm 1.5}$   & 37.7$_{\pm 0.9}$ \\
		r(2+1)d 152         & IG65M $\rightarrow$ K400      & 5.5$_{\pm 0.1}$     & 18.1$_{\pm 0.1}$    & 27.3$_{\pm 0.1}$    & 184.1$_{\pm 1.2}$   & 39.3$_{\pm 0.5}$ \\
		r(2+1)d 152         & Sports1M $\rightarrow$ K400   & 5.3$_{\pm 0.1}$     & 17.3$_{\pm 0.1}$    & 26.0$_{\pm 0.1}$    & 193.4$_{\pm 3.6}$   & 42.3$_{\pm 0.5}$ \\
		r(2+1)d 34          & IG65M			    & 9.1$_{\pm 0.2}$     & 27.2$_{\pm 0.2}$    & 38.7$_{\pm 0.2}$    & 108.1$_{\pm 0.0}$   & 19.0$_{\pm 0.0}$ \\ 
		r(2+1)d 34          & IG65M $\rightarrow$ K400	    & 8.2$_{\pm 0.2}$     & 25.3$_{\pm 0.3}$    & 36.7$_{\pm 0.1}$    & 120.8$_{\pm 0.7}$   & 21.0$_{\pm 0.0}$ \\
		CLIP	            & CLIP			    & \B{14.4$_{\pm 0.1}$} &  \B{37.4$_{\pm 0.3}$} & \B{50.2$_{\pm 0.3}$} & \B{70.3$_{\pm 0.3}$} & \B{10.3$_{\pm 0.5}$} \\
		s3dg MIL-NCE        & HowTo100M			    & 8.6$_{\pm 0.4}$      & 26.3$_{\pm 0.5}$      & 37.9$_{\pm 0.7}$     & 104.4$_{\pm 2.2}$    & 19.3$_{\pm 0.5}$ \\
		\bottomrule
	\end{tabular}	
	\caption{Comparison of the best available pretrain models as the motion experts for MMT.
			 IG65M $\rightarrow$ K400 means that model was trained on IG65M and then fine tuned on Kinetics400.
			 Results for each experiment are computed over three runs with random seeds. The results are reported on MSRVTT full clean split.}
	\label{tab:video_experts_fps32}
\end{table*}

As we can see, usually stronger models provide better results, but not always. Refer to r(2+1)d 152 rows, this network demonstrates one of the best
performance on Kinetics 400 benchmark, but works poorly as motion expert. Maybe this network is over specialized for Kinetics 400. More shallow analogue of r(2+1)d 152 is r(2+1)d 34 which shows much better results.

An interesting observation is that the best results are achieved with the networks trained in the unsupervised manner. CLIP and models trained on IG65M outperform all other models
trained on Kinetics in the supervised manner.
Another weakly supervised dataset is Sports1M~\cite{KarpathyCVPR14}. Models trained on this dataset provide
weak embeddings similar to the weak s3d model trained on Kinetics dataset.
The CLIP~\cite{radford2learning} (ViT-B/32) image feature extractor with a large margin outperforms all other models. The model s3dg MIL-NCE is a video encoder from the work~\cite{miech2020endtoend}. This network was trained from scratch on HowTo100M dataset.

As we show in Sec.~\ref{sec:tt_isect} Kinetics dataset has an overlap with MSRVTT dataset, and we don't know whether it affects to overfitting or not. Also it is worth to mention that IG65M and CLIP datasets are not publicly available, so we do not know if there is an overlap with MSRVTT and other video retrieval datasets.

For more details about our usage of pretrained video experts please refer to Sec.~\ref{sec:pretrain_experts}.

\begin{table*}[h]
  \centering
  \begin{tabular}{|l @{\hspace{1\tabcolsep}} |@{\hspace{1\tabcolsep}} l@{\hspace{1\tabcolsep}}l@{\hspace{1\tabcolsep}}l@{\hspace{1\tabcolsep}}l@{\hspace{1\tabcolsep}}l|}
    \toprule
    \multirow{2}{*}{model} & \multicolumn{5}{c|}{ActivityNet text $\rightarrow$ video} \\
                          & R@1$\uparrow$ & R@5$\uparrow$ & R@10$\uparrow$ & MnR$\downarrow$ & MdR$\downarrow$ \\
    \midrule
      CLIP~\cite{radford2learning}                 & 0.02 & 0.06 & 0.2 & 2210 & 2251 \\
      MMT (A$_\text{p/r}$) motion+audio~\cite{gabeur2020multimodal} & 7.3 & 22.5 & 31 & 283.9 & 30 \\
      \ours MDMMT(M$_c$ALVYMTS) L9H8 irCSN152+audio         & 15.1$_{\pm 0.1}$ & 38.3$_{\pm 0.1}$ & 51.5$_{\pm 0.3}$ & 92.4$_{\pm 2.3}$ & 10.0$_{\pm 0.0}$ \\
      \ours MDMMT(M$_c$ALVYMTS) L9H8 CLIP+audio             & 17.7$_{\pm 0.1}$ & 41.6$_{\pm 0.3}$ & 54.3$_{\pm 0.2}$ & 76.0$_{\pm 1.0}$ & 8.3$_{\pm 0.5}$ \\
      \ours MDMMT(M$_c$ALVYMTS) L9H8 CLIP+irCSN152+audio    & \B{20.1$_{\pm 0.5}$} & \B{45.1$_{\pm 0.5}$} & \B{58.0$_{\pm 0.6}$} & \B{70.8$_{\pm 0.1}$} & \B{7.0$_{\pm 0.0}$} \\
    \bottomrule
  \end{tabular}
  \caption{Test results on our split (see Sec.~\ref{sect:datasets}) on ActivityNet.}
  \label{tab:models-anet}
\end{table*}
\subsection{Datasets combination} \label{ssec:main}

In this section we show our experiments about the combination of different datasets. Nowadays video-caption datasets are not big enough to capture all real life
situations, also some datasets may be biased. The combination of different datasets may help to tackle this problem. 

Our experiments show that the proper combination of datasets allows to train a single model that can capture the knowledge from all used datasets. Important thing here is that in most cases the 
model trained on the combination of datasets is better than the model trained on a single dataset. 

In our experiments we combine all datasets presented in Tab.~\ref{tab:dataset-wgh}.
The important thing is how to sample minibatches during training. In our experiments we first sample a dataset, then we uniformly sample a video segment, if this sampled
video segment has more than one caption we sample a single caption uniformly. Column weight in Tab.~\ref{tab:dataset-wgh} describes the probability of sampling the corresponding dataset.
To obtain the probability of sampling the dataset with the weight $w$ we should divide $w$ by the sum of all weights.

The weights for all datasets are manually adjusted. It is important to find a good weight combination, because if some weight will be larger than needed, this dataset will be
overseen and as a result the performance will be lower comparing to the optimal case.
The opposite case is when a small weight was selected, this causes the situation when during training a network does not see the required number of 
examples from this dataset. 

For experiments in this section we use MMT with the only motion modality. 
Embeddings for the motion modality are computed with irCSN152 pretrained on IG65M.
All configurations are trained with 50 epochs and different number examples per epoch.
The initial learning rate is 5e-5. After each epoch we multiply learning rate by 0.95.
The MALVYMTS (see Tab.~\ref{tab:abbrev_names} for abbreviations.) configuration is trained with 150K examples per epoch. Configurations with the less number of datasets are trained with the less number of
examples per epoch.
The number of examples per epoch can be represented as a product of 150K by a sum of normalized weights (weights from Tab.~\ref{tab:dataset-wgh} divided by a sum of all weights) for each dataset (the initial sum equals to 1):
$150K = 150K \times (p_{\text{MSRVTT}} + p_{\text{ActivityNet}} + p_{\text{LSMDC}} + p_{\text{Twitter Vines}} + p_{\text{YouCook2}} + p_{\text{MSVD}} + p_{\text{TGIF}} + p_{\text{Something V2}})$.
If some dataset is removed from the training, we remove the corresponding coefficient from this sum, so the resulting length will be 150K multiplied by a value less than 1.

As far as we use the configurations M$_c$, A, L as the baselines, we need to be sure that the results for these configurations are the optimal values. 
In addition to the rule described above we try several values for a number of examples per epoch parameter, and report the results for the best found value.

\begin{table}[h]
      \centering
      \begin{tabular}{|c|c|}
	    \toprule
	    Dataset	    & Weight \\
	    \midrule
	    MSRVTT	    & 140 \\
	    ActivityNet	    & 100 \\
	    LSMDC	    & 70  \\
	    Twitter Vines   & 60  \\
	    YouCook2	    & 9   \\
	    MSVD	    & 9   \\
	    TGIF	    & 102 \\
	    Something V2    & 169 \\
	    \bottomrule
      \end{tabular}
      \caption{These datasets were used in our train procedure. The "Weight" column describes how often we sample examples from the dataset. The probability of obtaining an example from the
      dataset with the weight $w$ equals to $w$ divides by a sum of all weights.}
      \label{tab:dataset-wgh}
\end{table}

\begin{table}[h]
  \centering
  \begin{tabular}{|c|l @{\hspace{1\tabcolsep}} l @{\hspace{1\tabcolsep}} l|}
		  \toprule
	      \multirow{2}{*}{Dataset}  & \multicolumn{3}{c|}{Test Text $ \rightarrow$ Video  R@5 $ \uparrow$} \\
         &  MSRVTT & ActivityNet &       LSMDC \\
		  \midrule
    M$_c$        & 29.0$_{\pm0.2}$              & {\textit{13.4}$_{\pm0.3}$} &  {\textit{12.9}$_{\pm0.6}$} \\
    A        & {\textit{14.7}$_{\pm0.1}$} & 30.9$_{\pm0.6}$              &  {\textit{10.4}$_{\pm0.3}$} \\
    L        & {\textit{8.8}$_{\pm0.1}$}  & {\textit{7.2}$_{\pm0.2}$}  &  24.7$_{\pm0.6}$ \\
    M$_c$ALV     &  32.1$_{\pm0.1}$             & 32.0$_{\pm0.2}$              &  26.5$_{\pm0.7}$ \\
    M$_c$ALVYMT  &  33.8$_{\pm0.1}$             & 32.3$_{\pm0.2}$              &  27.3$_{\pm0.4}$ \\
    M$_c$ALVYMTS &  \textbf{34.5$_{\pm0.1}$}    & \textbf{32.4$_{\pm0.5}$}     &  \textbf{27.4$_{\pm0.6}$} \\ 
		  \bottomrule
	    \end{tabular}
  \caption{See abbreviations for the first column in Tab.~\ref{tab:abbrev_names}. The first three rows M$_c$,A,L report the quality of models trained on a single domain, and tested on other domains.
  \textit{Italic} means that the model did not see data from this domain during training. In this table the only motion modality (irCSN152) is used.}
  \label{tab:cross-dataset-test}
\end{table}

Tab.~\ref{tab:cross-dataset-test} summarizes our experiments on the datasets combination (for more details please refer to Sec.~\ref{sec:datasets_combination}). The main point here is that the proper combination of datasets leads to the best solution.

\subsection{Final result}
In this section we compare our solution with the prior art. Our two best solution uses three modalities: the audio, the motion and the RGB. To fuse modalities
we use MMT architecture with 9 layers and 8 heads. As a feature extractor for the audio stream the vggish~\cite{hershey2017cnn} network is used. For the video encoding we use
CLIP ViT-B/32 (RGB modality) and irCSN152 (motion modality) pretrained on IG65M dataset. The details about preprocessing videos for both networks are presented in Sec.~\ref{sec:pretrain_experts}.

Additionally we report separate results for motion + audio encoders and RGB + audio encoders because 
we do not know whether the IG65M or CLIP train database has a significant overlap with any of the test datasets or not.

All our models presented in Tab.~\ref{tab:models-anet},\ref{tab:models-msrvtt-full} and  \ref{tab:models-lsmdc}  are trained based on the pretrain HowTo100M model.
We present the details about pretraining in Sec.~\ref{sect:pretrain_ht100m}.

The results for MSRVTT are presented in Tab.~\ref{tab:models-msrvtt-full}. As we can see our solution MDMMT(MALVYMTS) L9H8 CLIP+irCSN152+audio
significantly outperforms all previous solutions on all splits: full, 1k-A and 1k-B. Our solution is better than the previous SotA (on R@5) on 8.7\%, 10.5\% and 14.4\% on
full, 1k-A and 1k-B correspondingly.
It is also worth to mention that our MDMMT (using only the motion, the RGB and the audio modalities) outperforms the original MMT (the motion, the RGB and the audio and 4 other modalities) by
8.7\%, 10.5\% and 14.4 (R@5) on full, 1k-A and 1k-B correspondingly.

We also report the results for the original CLIP~\cite{radford2learning}. The CLIP model has an image encoder and a text encoder, both pretrained in an unsupervised way. To test the CLIP model we take a single
frame from the middle of the video (this is the original testing protocol for CLIP). The row CLIP agg~\cite{portilloquintero2021straightforward} represents 
the usage of CLIP model with several frames using some specific aggregation procedure from this work.

In Tab.~\ref{tab:models-lsmdc} we report the results on LSMDC. On this benchmark we outperform the previous SotA solution by 8.6\%.

As we mention in Sec.~\ref{sect:datasets} we do not use the standard ActivityNet paragraph retrieval test protocol. Instead we use the text to video retrieval protocol.
To compare our solution with the previous work we take the previous SotA approach (MMT) in text to video retrieval and test it on our split.
The results are reported in Tab.~\ref{tab:models-anet}.
Our solution outperforms MMT by 22.6\%. The row MMT (A$_\text{p/r}$) motion+audio means that this network was trained only on ActivityNet dataset with the paragraph
retrieval mode. It is also worth to mention that CLIP shows very bad results on this benchmark. We try to aggregate with the mean pooling of 2, 4 and 16 uniformly taken embeddings, take the first 10, 20 and 70 words
from a caption, and no method improves the results.

The important property of our model is that we train a single model and test it on different test sets.
The authors of previous SotA approach (MMT) trained three different models for MSRVTT, ActivityNet and LSMDC, while in Tab.~\ref{tab:cross-dataset-test} we show that the model trained in such a manner
has poor generalization and can show good performance on the test part of the dataset $X$ if and only if it was trained on the train part of the dataset $X$.

\begin{table*}[h]
  \centering
  \vspace{-2em}\begin{tabular}{|l @{\hspace{1\tabcolsep}}  |@{\hspace{1\tabcolsep}} c @{\hspace{1\tabcolsep}} |@{\hspace{1\tabcolsep}} l @{\hspace{1\tabcolsep}} l @{\hspace{1\tabcolsep}} l @{\hspace{1\tabcolsep}} l @{\hspace{1\tabcolsep}} l|}
    \toprule
    \multirow{2}{*}{model} & \multirow{2}{*}{\rotatebox{90}{split}} & \multicolumn{5}{c|}{MSRVTT text $\rightarrow$ video} \\
			   &                        & R@1$\uparrow$ & R@5$\uparrow$ & R@10$\uparrow$ & MnR$\downarrow$ & MdR$\downarrow$ \\
    \midrule
      Random baseline &\multirow{14}{*}{\rotatebox{90}{full}} & 0.0 & 0.2 & 0.3 & 1500 & 1500 \\
      VSE~\cite{mithun2018learning}                         && 5.0 & 16.4 & 24.6 & --- & 47 \\
      VSE++~\cite{mithun2018learning}                       && 5.7 & 17.1 & 24.8 & --- & 65 \\
      Multi Cues~\cite{mithun2018learning}                  && 7.0 & 20.9 & 29.7 & --- & 38 \\
      W2VV~\cite{Dong_2018}				    && 6.1 & 18.7 & 27.5 & --- & 45 \\
      Dual Enc.~\cite{dong2019dual}                         && 7.7 & 22.0 & 31.8 & --- & 32 \\
      CE~\cite{liu2020use}				    && 10.0$_{\pm 0.1}$ & 29.0$_{\pm 0.3}$ & 41.2$_{\pm 0.2}$ & 86.8$_{\pm 0.3}$ & 16.0$_{\pm 0.0}$ \\
      MMT (M) 7mod~\cite{gabeur2020multimodal}              && 10.7$_{\pm 0.2}$ &  31.1$_{\pm 0.1}$ & 43.4$_{\pm 0.2}$ &  88.2$_{\pm 0.7}$ & 15.0$_{\pm 0.0}$  \\
      CLIP~\cite{radford2learning}                 && 15.1 & 31.8 & 40.4 &  184.2  & 21 \\
      CLIP agg~\cite{portilloquintero2021straightforward}   && 21.5 & 41.1 & 50.4 &  --- & \B{4} \\
      \ours MDMMT(MALVYMTS) L9H8 irCSN152+audio             && 15.7$_{\pm 0.1}$ & 38.8$_{\pm 0.1}$ & 51.1$_{\pm 0.2}$ & 76.0$_{\pm 0.7}$ & 10.0$_{\pm 0.0}$ \\
      \ours MDMMT(MALVYMTS) L9H8 CLIP+audio                 && 21.7$_{\pm 0.2}$ & 47.6$_{\pm 0.3}$ & 59.8$_{\pm 0.1}$ & 55.9$_{\pm 0.2}$ & 6.0$_{\pm 0.0}$ \\
      \ours MDMMT(MALVYMTS) L9H8 CLIP+irCSN152+audio        && \B{23.1$_{\pm 0.1}$} & \B{49.8$_{\pm 0.1}$} & \B{61.8$_{\pm 0.1}$} & \B{52.8$_{\pm 0.2}$} & 6.0$_{\pm 0.0}$ \\
    \midrule
      MMT (M$_c$) 7mod~\cite{gabeur2020multimodal} &\multirow{4}{*}{\rotatebox{90}{full}\hspace{0.1em}\rotatebox{90}{clean}}& 10.4$_{\pm 0.1}$ & 30.2$_{\pm 0.4}$ & 42.3$_{\pm 0.2}$ & 89.4$_{\pm 0.6}$ & 15.7$_{\pm 0.5}$ \\
      \ours MDMMT(M$_c$ALVYMTS) L9H8 irCSN152+audio      && 15.8$_{\pm 0.1}$ & 38.9$_{\pm 0.1}$ & 51.0$_{\pm 0.1}$ & 76.4$_{\pm 0.5}$ & 10.0$_{\pm 0.0}$ \\
      \ours MDMMT(M$_c$ALVYMTS) L9H8 CLIP+audio && 21.5$_{\pm 0.1}$ & 47.4$_{\pm 0.2}$ & 59.6$_{\pm 0.1}$ & 57.7$_{\pm 0.4}$ & 6.0$_{\pm 0.0}$ \\
      \ours MDMMT(M$_c$ALVYMTS) L9H8 CLIP+irCSN152+audio && \B{22.8$_{\pm 0.2}$} & \B{49.5$_{\pm 0.1}$} & \B{61.5$_{\pm 0.1}$} & \B{53.8$_{\pm 0.3}$} & \B{6.0$_{\pm 0.0}$} \\
    \midrule
      Random baseline &\multirow{12}{*}{\rotatebox{90}{1k-A}} &  0.1 & 0.5 & 1.0 & 500.0 & 500.0\\
      JSFusion~\cite{yu2018joint}		                  && 10.2 & 31.2 & 43.2 & --- & 13 \\
      E2E~\cite{miech2020endtoend}                                && 9.9 & 24.0 & 32.4 & --- & 29.5 \\
      HT~\cite{miech19howto100m}		                  && 14.9& 40.2 & 52.8 & --- & 9 \\
      CE~\cite{liu2020use}                                        && 20.9$_{\pm 1.2}$ & 48.8$_{\pm 0.6}$ & 62.4$_{\pm 0.8}$ & 28.2$_{\pm 0.8}$ & 6.0$_{\pm 0.0}$ \\
      CLIP~\cite{radford2learning}	                  && 22.5 & 44.3 & 53.7 & 61.7 & 8 \\
      MMT (M$_\text{1k-A}$) 7mod~\cite{gabeur2020multimodal} && 26.6$_{\pm 1.0}$   & 57.1$_{\pm 1.0}$   & 69.6$_{\pm 0.2}$   & 24.0$_{\pm 0.8}$   & 4.0$_{\pm 0.0}$ \\
      AVLnet\cite{rouditchenko2020avlnet}                         && 27.1 & 55.6 & 66.6 & --- & 4\\
      SSB~\cite{patrick2021supportset}		                  && 30.1 & 58.5 & 69.3 & --- & 3.0 \\
      CLIP agg~\cite{portilloquintero2021straightforward}         && 31.2 & 53.7 & 64.2 & --- & 4 \\
      \ours MDMMT(M$_{\text{1k-A}}$ALVYMTS) L9H8 irCSN152+audio   && 31.3$_{\pm 0.1}$ & 60.4$_{\pm 1.2}$ & 71.8$_{\pm 1.0}$ & 24.0$_{\pm 0.4}$ & 3.0$_{\pm 0.0}$ \\
      \ours MDMMT(M$_{\text{1k-A}}$ALVYMTS) L9H8 CLIP+audio        && 38.9$_{\pm 1.0}$	& 68.3$_{\pm 0.7}$ & 78.8$_{\pm 0.2}$ & 17.3$_{\pm 0.5}$ & \B{2.0$_{\pm 0.0}$} \\
      \ours MDMMT(M$_{\text{1k-A}}$ALVYMTS) L9H8 CLIP+irCSN152+audio && \B{38.9$_{\pm 0.6}$} & \B{69.0$_{\pm 0.1}$} & \B{79.7$_{\pm 0.6}$} & \B{16.5$_{\pm 0.4}$} & \B{2.0$_{\pm 0.0}$} \\
    \midrule
      Random baseline &\multirow{10}{*}{\rotatebox{90}{1k-B}} &  0.1 & 0.5 & 1.0 & 500.0 & 500.0\\
      MEE~\cite{miech2020learning} && 13.6 & 37.9 & 51.0 & --- & 10.0 \\
      JPose~\cite{wray2019finegrained} && 14.3 & 38.1 & 53.0 & --- & 9 \\
      MEE-COCO~\cite{miech2020learning} && 14.2 & 39.2 & 53.8 & --- & 9.0 \\
      CE~\cite{liu2020use} && 18.2$_{\pm 0.7}$ & 46.0$_{\pm 0.4}$ & 60.7$_{\pm 0.2}$  & 35.3$_{\pm 1.1}$ & 7.0$_{\pm 0.0}$  \\
      MMT (M$_\text{1k-B}$) 7mod~\cite{gabeur2020multimodal} &&  24.5$_{\pm0.5}$  &  54.4$_{\pm0.8}$	& 68.0$_{\pm0.5}$  & 26.6$_{\pm0.2}$ & 4.7$_{\pm0.5}$ \\
      CLIP~\cite{radford2learning}					&& 24.5 & 46.2 & 56.8 & 60.9 & 7 \\
      \ours MDMMT(M$_{\text{1k-B}}$ALVYMTS) L9H8 irCSN152+audio&& 28.8$_{\pm 0.9}$ & 58.8$_{\pm 0.3}$ & 71.2$_{\pm 0.3}$ & 28.5$_{\pm 0.5}$ & 3.7$_{\pm 0.5}$\\
      \ours MDMMT(M$_{\text{1k-B}}$ALVYMTS) L9H8 CLIP+audio    &&  35.1$_{\pm 0.1}$ &  66.5$_{\pm 0.9}$ & 77.6$_{\pm 0.3}$ & 21.5$_{\pm 0.4}$ & 2.7$_{\pm 0.5}$ \\
      \ours MDMMT(M$_{\text{1k-B}}$ALVYMTS) L9H8 CLIP+irCSN152+audio       && \B{37.4$_{\pm 1.5}$} & \B{68.8$_{\pm 0.4}$} & \B{79.4$_{\pm 0.4}$} & \B{21.3$_{\pm 0.4}$} & \B{2.0$_{\pm 0.0}$} \\

    \bottomrule
  \end{tabular}
  \caption{Results on MSRVTT dataset.}
  \label{tab:models-msrvtt-full}

  \vspace{\floatsep}

  \begin{tabular}{|l@{\hspace{1\tabcolsep}} |l@{\hspace{1\tabcolsep}}l@{\hspace{1\tabcolsep}}l@{\hspace{1\tabcolsep}}l@{\hspace{1\tabcolsep}}l@{\hspace{1\tabcolsep}}|l|}
    \toprule
    \multirow{2}{*}{model} & \multicolumn{5}{c|}{LSMDC text $\rightarrow$ video} \\
			   & R@1$\uparrow$ & R@5$\uparrow$ & R@10$\uparrow$ & MnR$\downarrow$ & MdR$\downarrow$ \\
    \midrule
      CT-SAN~\cite{yu2017endtoend}	                  & 5.1 & 16.3  & 25.2 & --- & 46 \\
      JSFusion~\cite{yu2018joint}		          & 9.1 & 21.2  & 34.1 & --- & 36 \\
      MEE~\cite{miech2020learning}                        & 9.3 & 25.1  & 33.4 & --- & 27 \\
      MEE-COCO~\cite{miech2020learning}                   & 10.1 & 25.6 & 34.6 & --- & 27 \\
      CE~\cite{liu2020use}                                & 11.2$_{\pm 0.4}$ & 26.9$_{\pm 1.1}$ & 34.8$_{\pm 2.0}$ & 96.8$_{\pm 5.0}$ & 25.3$_{\pm 3.1}$ \\
      CLIP agg~\cite{portilloquintero2021straightforward} & 11.3 & 22.7 & 29.2 & --- & 56.5 \\
      CLIP~\cite{radford2learning}               & 12.4 & 23.7 & 31.0  & 142.5 & 45  \\
      MMT (L) 7mod~\cite{gabeur2020multimodal}            & 12.9$_{\pm 0.1}$ & 29.9$_{\pm 0.7}$ & 40.1$_{\pm 0.8}$ & 75.0$_{\pm 1.2}$ & 19.3$_{\pm 0.2}$ \\
      \ours MDMMT(M$_c$ALVYMTS) L9H8 irCSN152+audio       & 13.1$_{\pm 0.5}$ & 31.3$_{\pm 0.3}$ & 40.1$_{\pm 0.0}$ & 74.5$_{\pm 0.7}$ & 19.3$_{\pm 0.5}$ \\
      \ours MDMMT(M$_c$ALVYMTS) L9H8 CLIP+audio           & 17.2$_{\pm 0.6}$ & 34.9$_{\pm 0.4}$ & 45.3$_{\pm 1.0}$ & 65.6$_{\pm 0.8}$ & 14.0$_{\pm 0.8}$ \\
      \ours MDMMT(M$_c$ALVYMTS) L9H8 CLIP+irCSN152+audio  & \B{18.8$_{\pm 0.7}$} & \B{38.5$_{\pm 0.4}$} & \B{47.9$_{\pm 0.7}$} & \B{58.0$_{\pm 1.1}$} & \B{12.3$_{\pm 0.5}$} \\
    \bottomrule
  \end{tabular}
  \caption{Test results on LSMDC public test (1k video)}
  \label{tab:models-lsmdc}
\end{table*}

\section{Conclusions and Discussion}
In this work we present a new text to video retrieval state-of-the-art model on MSRVTT and LSMDC benchmarks. We do not use ActivityNet dataset in the paragraph retrieval mode
as many previous works do, so we can't compare with them. But we show that on ActivityNet in the video retrieval mode we outperform the previous state-of-the-art model (MMT) by a large margin.
Our model has captured knowledge from many video caption datasets, thus it is able to show the best results on several datasets at the same time without finetuning.

We also present a practical approach to find the overlap between two different video datasets. Using this approach we find the overlap between several datasets.
Especially we find a large overlap between the MSRVTT test and train parts, and between the ActivityNet test and train parts. Removing this overlap from the MSRVTT train part significantly
decreases the performance of previous best models on MSRVTT benchmark.

\textit{Acknowledgments.} We would like to thank Andrey Ivanyuta and other colleagues from Intelligent Systems and Data Science Lab for helping to find the overlap between datasets.

\clearpage

\printbibliography



\appendix

\section{Pretrain experts usage} \label{sec:pretrain_experts}
The important data preparing stage is how to sample frames from a video to achieve the best performance.
For s3d experiments the input video is converted to 30 frames per second, for all other experiments we convert
the input video to 32 frames per second. As a result we compute a single embedding for each second, having 1 second window with 1 second shift (no overlapping).

The input frame size is important. We use the different sizes for the different models. For each model we use the recommended input size.
For s3d we resize a video to 256 on the short side and then take a 224x224 center crop.
For SlowFast 32x2 R101 we resize a video to 256 on the short side and then take a 256x256 center crop.
For ipCSN 152 and irCSN 152 we resize a video to 224 on the short side and take a 224x224 center crop.
For r(2+1)d 152 and r(2+1)d 34 we resize a video to 112 on the short side and then take a 112x112 center crop.

Pretrained models for ipCSN, irCSN and r(2+1)d are available here\footnote{https://github.com/facebookresearch/VMZ},
for SlowFast 32x2 R101 here\footnote{https://github.com/facebookresearch/SlowFast/blob/master/MODEL\_ZOO.md}, and for s3d here\footnote{https://github.com/princeton-vl/d3dhelper/blob/master/d3d\_helper.ipynb}.

For the CLIP model~\cite{radford2learning} we resize a video to 224 on the short side and take a center crop, then we extract 1 frame per second.
We use a publicly available image encoder. We do not use the text encoder from CLIP.

Model s3dg MIL-NCE is a video encoder from the work~\cite{miech2020endtoend}. This network was trained from scratch on HowTo100M dataset. For this network we resize
the input video stream to the size of 228x228 pixels, then take a center crop.

\section{Datasets combination} \label{sec:datasets_combination}
\begin{figure}[h]
      \centering
\scalebox{0.90}{
\begin{tikzpicture}
  \draw[->,ultra thick] (0,0)--(0.9\columnwidth,0);
  \draw[->,ultra thick] (0,0)--(0,0.8166666666666667\columnwidth);
  \node[align=left] at (0.19\columnwidth,0.7166666666666667\columnwidth) {LSMDC R@5$\uparrow$};
  \node [draw,shape=circle,fill=black,minimum size=0.2cm,inner sep=0pt] at (0.1\columnwidth,0){};
  \node[align=center] at (0.1\columnwidth,-0.05\columnwidth) {L};
  \node [draw,shape=circle,fill=black,minimum size=0.2cm,inner sep=0pt] at (0.1\columnwidth,0.1\columnwidth){};
  \node[align=right] at (-0.07\columnwidth,0.1\columnwidth) {24.7};
  \draw[loosely dotted] (0.1\columnwidth,0.1\columnwidth) -- (0.1\columnwidth,0\columnwidth);
  \draw[loosely dotted] (0.1\columnwidth,0.1\columnwidth) -- (0\columnwidth,0.1\columnwidth);
  \node [draw,shape=circle,fill=black,minimum size=0.2cm,inner sep=0pt] at (0.25\columnwidth,0){};
  \node[align=center] at (0.25\columnwidth,-0.05\columnwidth) {M$_c$ALV};
  \node [draw,shape=circle,fill=black,minimum size=0.2cm,inner sep=0pt] at (0.25\columnwidth,0.4663595833125108\columnwidth){};
  \node[align=right] at (-0.07\columnwidth,0.4663595833125108\columnwidth) {26.5};
  \draw[loosely dotted] (0.25\columnwidth,0.4663595833125108\columnwidth) -- (0.25\columnwidth,0\columnwidth);
  \draw[loosely dotted] (0.25\columnwidth,0.4663595833125108\columnwidth) -- (0\columnwidth,0.4663595833125108\columnwidth);
  \node [draw,shape=circle,fill=black,minimum size=0.2cm,inner sep=0pt] at (0.5\columnwidth,0){};
  \node[align=center] at (0.5\columnwidth,-0.05\columnwidth) {M$_c$ALVYMT};
  \node [draw,shape=circle,fill=black,minimum size=0.2cm,inner sep=0pt] at (0.5\columnwidth,0.6459452442507194\columnwidth){};
  \draw[loosely dotted] (0.5\columnwidth,0.6459452442507194\columnwidth) -- (0.5\columnwidth,0\columnwidth);
  \node [draw,shape=circle,fill=black,minimum size=0.2cm,inner sep=0pt] at (0.8\columnwidth,0){};
  \node[align=center] at (0.8\columnwidth,-0.05\columnwidth) {M$_c$ALVYMTS};
  \node [draw,shape=circle,fill=black,minimum size=0.2cm,inner sep=0pt] at (0.8\columnwidth,0.6666666666666666\columnwidth){};
  \node[align=right] at (-0.07\columnwidth,0.6666666666666666\columnwidth) {27.4};
  \draw[loosely dotted] (0.8\columnwidth,0.6666666666666666\columnwidth) -- (0.8\columnwidth,0\columnwidth);
  \draw[loosely dotted] (0.8\columnwidth,0.6666666666666666\columnwidth) -- (0\columnwidth,0.6666666666666666\columnwidth);
  \draw[thick,color=black!60!green] (0.1\columnwidth,0.1\columnwidth) -- (0.25\columnwidth,0.4663595833125108\columnwidth);
  \draw[thick,color=black!60!green] (0.25\columnwidth,0.4663595833125108\columnwidth) -- (0.5\columnwidth,0.6459452442507194\columnwidth);
  \draw[thick,color=black!60!green] (0.5\columnwidth,0.6459452442507194\columnwidth) -- (0.8\columnwidth,0.6666666666666666\columnwidth);
\end{tikzpicture}
}
      \caption{Increasing R@5 metric on the LSMDC test set while enriching the train part.}
      \label{fig:lsmdc-dataset-combine}
\end{figure}
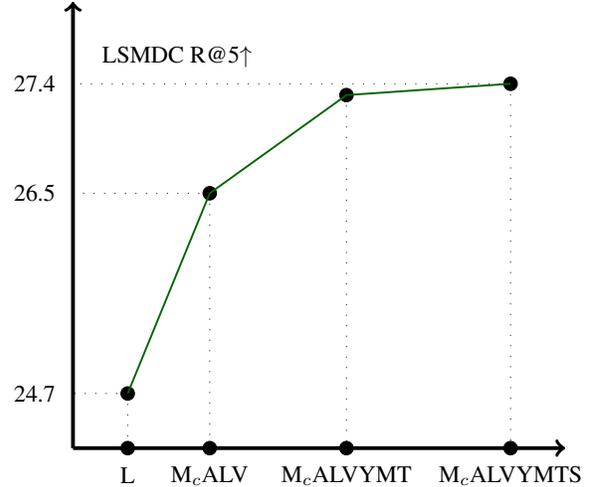

In Fig.~\ref{fig:msrvtt-dataset-combine},\ref{fig:activitynet-dataset-combine},\ref{fig:lsmdc-dataset-combine} we present 6 models.
Abbreviations M$_c$ALV, M$_c$ALVYMTS and M$_c$ALVYMTS represent the same three models on these figures.
The first model, called M$_c$, is trained on the MSRVTT full clean split only,
the second one, called A, is trained on ActivityNet only. And the third model, called L, is trained on LSMDC only.
These three models are taken as baselines. Adding more datasets 
should be not worse than these baseline. The forth model is called M$_c$ALV. This model is trained on
the combination of MSRVTT, ActivityNet, LSMDC and TwitterVines. As we can see M$_c$$\rightarrow$M$_c$ALV gives +3.07\% on MSRVTT (full clean split), A$\rightarrow$M$_c$ALV gives +1.06\% on ActivityNet,
and L$\rightarrow$M$_c$ALV gives +1.77\% on LSMDC. The next model is called M$_c$ALVYMT and it is trained on combination of MSRVTT, ActivityNet, LSMDC, TwitterVines, YouCook2, MSVD, TGIF.
The transitions M$_c$$\rightarrow$M$_c$ALVYMT, A$\rightarrow$M$_c$ALVYMT, L$\rightarrow$M$_c$ALVYMT give +4.85\%, +1.45\% and +2.63\% correspondingly. The last transitions M$_c$$\rightarrow$M$_c$ALVYMTS,
A$\rightarrow$M$_c$ALVYMTS, L$\rightarrow$M$_c$ALVYMTS slightly improve the performance on ActivityNet and LSMDC and significantly improve the performance on MSRVTT.
Finally, the combination of all datasets gives +5.5\% for MSRVTT, +1.47\% for ActivityNet and +2.74\% for LSMDC.

\begin{figure}[h]
      \centering
\scalebox{0.90}{
\begin{tikzpicture}
  \draw[->,ultra thick] (0,0)--(0.9\columnwidth,0);
  \draw[->,ultra thick] (0,0)--(0,0.8166666666666667\columnwidth);
  \node[align=left] at (0.3\columnwidth,0.7166666666666667\columnwidth) {MSRVTT full clean R@5 $\uparrow$};
  \node [draw,shape=circle,fill=black,minimum size=0.2cm,inner sep=0pt] at (0.1\columnwidth,0){};
  \node[align=center] at (0.1\columnwidth,-0.05\columnwidth) {M$_c$};
  \node [draw,shape=circle,fill=black,minimum size=0.2cm,inner sep=0pt] at (0.1\columnwidth,0.1\columnwidth){};
  \node[align=right] at (-0.07\columnwidth,0.1\columnwidth) {29.0};
  \draw[loosely dotted] (0.1\columnwidth,0.1\columnwidth) -- (0.1\columnwidth,0\columnwidth);
  \draw[loosely dotted] (0.1\columnwidth,0.1\columnwidth) -- (0\columnwidth,0.1\columnwidth);
  \node [draw,shape=circle,fill=black,minimum size=0.2cm,inner sep=0pt] at (0.25\columnwidth,0){};
  \node[align=center] at (0.25\columnwidth,-0.05\columnwidth) {M$_c$ALV};
  \node [draw,shape=circle,fill=black,minimum size=0.2cm,inner sep=0pt] at (0.25\columnwidth,0.41658278490608236\columnwidth){};
  \node[align=right] at (-0.07\columnwidth,0.41658278490608236\columnwidth) {32.1};
  \draw[loosely dotted] (0.25\columnwidth,0.41658278490608236\columnwidth) -- (0.25\columnwidth,0\columnwidth);
  \draw[loosely dotted] (0.25\columnwidth,0.41658278490608236\columnwidth) -- (0\columnwidth,0.41658278490608236\columnwidth);
  \node [draw,shape=circle,fill=black,minimum size=0.2cm,inner sep=0pt] at (0.5\columnwidth,0){};
  \node[align=center] at (0.5\columnwidth,-0.05\columnwidth) {M$_c$ALVYMT};
  \node [draw,shape=circle,fill=black,minimum size=0.2cm,inner sep=0pt] at (0.5\columnwidth,0.599237741699757\columnwidth){};
  \node[align=right] at (-0.07\columnwidth,0.599237741699757\columnwidth) {33.8};
  \draw[loosely dotted] (0.5\columnwidth,0.599237741699757\columnwidth) -- (0.5\columnwidth,0\columnwidth);
  \draw[loosely dotted] (0.5\columnwidth,0.599237741699757\columnwidth) -- (0\columnwidth,0.599237741699757\columnwidth);
  \node [draw,shape=circle,fill=black,minimum size=0.2cm,inner sep=0pt] at (0.8\columnwidth,0){};
  \node[align=center] at (0.8\columnwidth,-0.05\columnwidth) {M$_c$ALVYMTS};
  \node [draw,shape=circle,fill=black,minimum size=0.2cm,inner sep=0pt] at (0.8\columnwidth,0.6666666666666666\columnwidth){};
  \node[align=right] at (-0.07\columnwidth,0.6666666666666666\columnwidth) {34.5};
  \draw[loosely dotted] (0.8\columnwidth,0.6666666666666666\columnwidth) -- (0.8\columnwidth,0\columnwidth);
  \draw[loosely dotted] (0.8\columnwidth,0.6666666666666666\columnwidth) -- (0\columnwidth,0.6666666666666666\columnwidth);
  \draw[thick,color=black!60!green] (0.1\columnwidth,0.1\columnwidth) -- (0.25\columnwidth,0.41658278490608236\columnwidth);
  \draw[thick,color=black!60!green] (0.25\columnwidth,0.41658278490608236\columnwidth) -- (0.5\columnwidth,0.599237741699757\columnwidth);
  \draw[thick,color=black!60!green] (0.5\columnwidth,0.599237741699757\columnwidth) -- (0.8\columnwidth,0.6666666666666666\columnwidth);
\end{tikzpicture}
}
  \caption{Increasing R@5 metric on the MSRVTT full clean split while enriching the train part.}
  \label{fig:msrvtt-dataset-combine}
\end{figure}
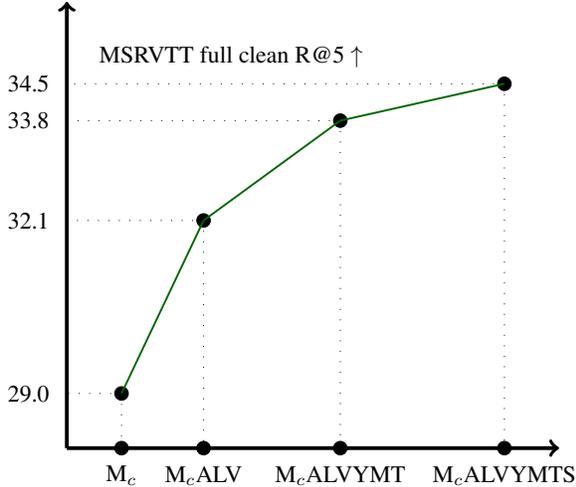

\begin{figure}[h]
      \centering
\scalebox{0.90}{
\begin{tikzpicture}
  \draw[->,ultra thick] (0,0)--(0.9\columnwidth,0);
  \draw[->,ultra thick] (0,0)--(0,0.8166666666666667\columnwidth);
  \node[align=left] at (0.19\columnwidth,0.7166666666666667\columnwidth) {ActivityNet R@5 $\uparrow$};
  \node [draw,shape=circle,fill=black,minimum size=0.2cm,inner sep=0pt] at (0.1\columnwidth,0){};
  \node[align=center] at (0.1\columnwidth,-0.05\columnwidth) {A};
  \node [draw,shape=circle,fill=black,minimum size=0.2cm,inner sep=0pt] at (0.1\columnwidth,0.1\columnwidth){};
  \node[align=right] at (-0.07\columnwidth,0.1\columnwidth) {30.9};
  \draw[loosely dotted] (0.1\columnwidth,0.1\columnwidth) -- (0.1\columnwidth,0\columnwidth);
  \draw[loosely dotted] (0.1\columnwidth,0.1\columnwidth) -- (0\columnwidth,0.1\columnwidth);
  \node [draw,shape=circle,fill=black,minimum size=0.2cm,inner sep=0pt] at (0.25\columnwidth,0){};
  \node[align=center] at (0.25\columnwidth,-0.05\columnwidth) {M$_c$ALV};
  \node [draw,shape=circle,fill=black,minimum size=0.2cm,inner sep=0pt] at (0.25\columnwidth,0.5076325287125595\columnwidth){};
  \node[align=right] at (-0.07\columnwidth,0.5076325287125595\columnwidth) {32.0};
  \draw[loosely dotted] (0.25\columnwidth,0.5076325287125595\columnwidth) -- (0.25\columnwidth,0\columnwidth);
  \draw[loosely dotted] (0.25\columnwidth,0.5076325287125595\columnwidth) -- (0\columnwidth,0.5076325287125595\columnwidth);
  \node [draw,shape=circle,fill=black,minimum size=0.2cm,inner sep=0pt] at (0.5\columnwidth,0){};
  \node[align=center] at (0.5\columnwidth,-0.05\columnwidth) {M$_c$ALVYMT};
  \node [draw,shape=circle,fill=black,minimum size=0.2cm,inner sep=0pt] at (0.5\columnwidth,0.6612756789394085\columnwidth){};
  \draw[loosely dotted] (0.5\columnwidth,0.6612756789394085\columnwidth) -- (0.5\columnwidth,0\columnwidth);
  \node [draw,shape=circle,fill=black,minimum size=0.2cm,inner sep=0pt] at (0.8\columnwidth,0){};
  \node[align=center] at (0.8\columnwidth,-0.05\columnwidth) {M$_c$ALVYMTS};
  \node [draw,shape=circle,fill=black,minimum size=0.2cm,inner sep=0pt] at (0.8\columnwidth,0.6666666666666666\columnwidth){};
  \node[align=right] at (-0.07\columnwidth,0.6666666666666666\columnwidth) {32.4};
  \draw[loosely dotted] (0.8\columnwidth,0.6666666666666666\columnwidth) -- (0.8\columnwidth,0\columnwidth);
  \draw[loosely dotted] (0.8\columnwidth,0.6666666666666666\columnwidth) -- (0\columnwidth,0.6666666666666666\columnwidth);
  \draw[thick,color=black!60!green] (0.1\columnwidth,0.1\columnwidth) -- (0.25\columnwidth,0.5076325287125595\columnwidth);
  \draw[thick,color=black!60!green] (0.25\columnwidth,0.5076325287125595\columnwidth) -- (0.5\columnwidth,0.6612756789394085\columnwidth);
  \draw[thick,color=black!60!green] (0.5\columnwidth,0.6612756789394085\columnwidth) -- (0.8\columnwidth,0.6666666666666666\columnwidth);
\end{tikzpicture}
}
      \caption{Increasing R@5 metric on the ActivityNet test set while enriching the train part.}
      \label{fig:activitynet-dataset-combine}
\end{figure}
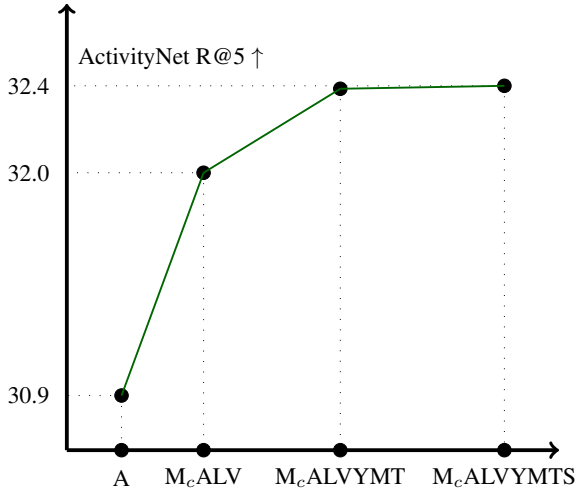

\section{Test and train intersection} \label{sec:tt_isect}

In this section we present our analysis of overlapping of popular text to video datasets. 
Since we compose the train dataset from several different datasets it is important to be sure that there is no the same video segment
in the train part and in the test part. Our aim is to find the overlap between the train part of used datasets~---
MSRVTT, ActivityNet, LSMDC, YouCook2, MSVD, TGIF, TwitterVines, HowTo100M, Kinetics700
and the test parts of MSRVTT, ActivityNet and LSMDC, and then to remove found duplicates from the train parts.

Note that for training  we use Something to Something V2 dataset, but we do not try to find overlap between it and test datasets
because this dataset is artificially created, thus the probability to find duplicates is very low.

We decided to find the overlap only for MSRVTT, ActivityNet and LSMDC because these are the most popular datasets and we do not have enough human resources
to find the overlap for the test part of all other datasets.

Our cleaning method consists of two stages. The first stage is to match video segments by the YouTube ID (if the ID is available) 
and remove from train parts all video segments that have the corresponding pair in test parts. In Tab.~\ref{tab:dataset_size}
the information about the availability of YouTube IDs in datasets is presented.
We collect the YouTube ID for all videos from MSRVTT full test and ActivityNet validation 1,2 and remove corresponding video segments from the train part. 

The second stage is based on matching frames by embeddings. For each video we compute several embeddings then we compute the similarity between each video from the train part and the test part. After we manually assess several thousands of video segments with highest scores
for each pair of datasets. Then we extend found duplicates by either the YouTube ID or the internal dataset ID.
This means that if a video $V_1$ is marked as a duplicate and a video $V_2$ is not marked as a duplicate, but they have the same YouTube ID or same internal dataset ID,
we will remove $V_1$ and $V_2$ from the train part.
In case of LSMDC we do not have the YouTube ID, but have the name of the movie from which the video segment was taken, so if a video segment $V_1$ is marked
as a duplicate, we remove all segments taken from the movie of $V_1$.
The detailed description of the second stage is described in Sec.~\ref{ssec:NDVS}.

Surprisingly we found that the MSRVTT test has a significant overlap with the MSRVTT train part. This problem is relevant for the full, 1k-A and 1k-B splits.
The ActivityNet dataset suffers from the same problem.

For large datasets like HowTo100M and Kinetics700 we can not find the whole intersection, but we estimate the approximate number of videos in the intersection.
We found that HowTo100M may have about 300 (10\% of the MSRVTT full test part) video segments that can be in the MSRVTT full test part.

The similar situation is about Kinetics700 and ActivityNet datasets. Kinetics700 may have approximately 500-600 video segments (10\% of the ActivityNet test)
that may have duplicates in ActivityNet validation 1,2. Another problem with the Kinetics dataset is that many motion models are pretrained on it.

This circumstance means that researchers should carefully use HowTo100M and Kinetics700 along with MSRVTT and ActivityNet correspondingly,
because for today we don't know whether a neural network overfits for some portion of this intersection or not.

\begin{figure*}[h!]
	\centering
	\includegraphics[width=13cm]{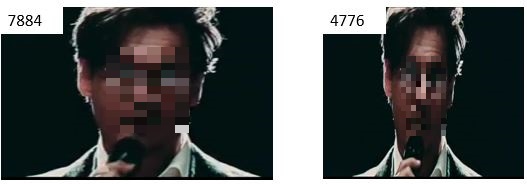} 
	\includegraphics[width=13cm]{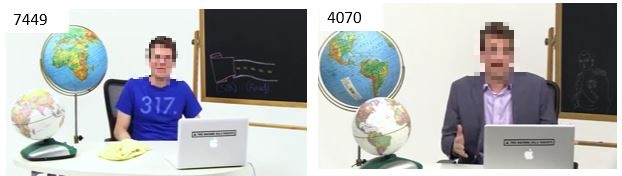} 
	\includegraphics[width=13cm]{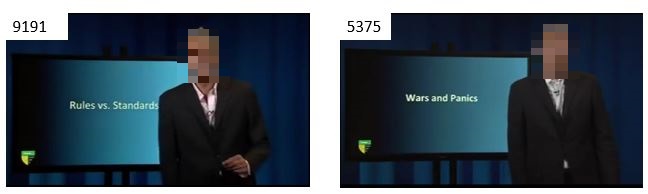} 
	\includegraphics[width=13cm]{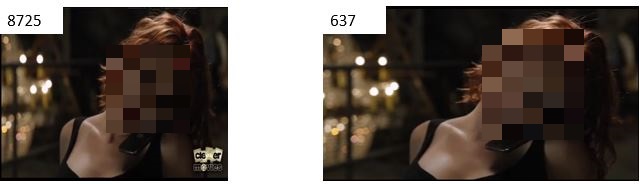} 
	
	\caption{The left image is taken from the MSRVTT test split and the right one from MSRVTT Train. The numbers in the upper left corner represent the MSRVTT video ID.
	The faces are blurred in order to avoid legal claims.}
	\label{fig:msrvtt_dubl}
\end{figure*}

All duplicates can be considered as two groups of pairs. 
Pairs from the first group have the same videos, but different brightness, aspect ratio, size, presence/absence of a logo and so on.
The second group has pairs with quite similar videos, for example it can be the same person on the same background, doing the same things, but wearing different clothes.
We think that it is better to remove such videos from the train part to prevent overfitting. Several found examples are presented in Fig.~\ref{fig:msrvtt_dubl}.

\subsection{Near duplicate video search} \label{ssec:NDVS}

\subsubsection{Approach}

In this section we explain our approach that is used to find the same or quite similar video segments in test and train parts.

Suppose we have two sets of videos $Q = \{q_1, ...., q_k\}$ and $G = \{g_1, ..., g_n\}$ called the query set and the gallery set.
We want to find all pairs $(q_i, g_j)$ where $q_i$ and  $g_j$ have a common video segment. 

From each $q_i$ and $g_j$ we extract 1 frame per second. Each video is then represented by a sequence of pictures:
$q_i = [q_i^1, ...., q_i^{s_i}]$ and $g_j = [g_j^1, ..., g_j^{p_j}]$. Then a 2D pretrained neural network is used
to extract features from each image: $\bar{q}_a^b = \texttt{neuralnet}(q_a^b)$ and $\bar{g}_a^b = \texttt{neuralnet}(g_a^b)$.

Then we compute the matrix of cosines between the features from Q and G: $s_{ij}^{ab} = \frac{<\bar{q}_i^a, \bar{g}_j^b>}{||\bar{q}_a^b||_2 ||\bar{g}_a^b||_2}$.

Now each pair $(q_i, g_j)$ is represented by the matrix:

\[
	\begin{array}{c|ccc}
			      & g_j^1         & ... & g_j^{p_j} \\ \hline
		q_i^1     & s_{ij}^{11}   & ... & s_{ij}^{1p_j} \\
		... \\
		q_i^{s_i} & s_{ij}^{s_i1} & ... & s_{ij}^{s_ip_j} \\
	\end{array}
\]

Suppose that videos $q_i$ and $g_j$ are intersected at time moments $t_q$ and $t_g$, it is naturally to assume that
the next several seconds $t_q+1, ..., t_q+K-1$ and $t_g+1, ..., t_g+K-1$ ($K \le \min(s_i, p_j)$) represent the same video segment.
Motivated by this fact we compute the mean cosine for each interval of K seconds (we use K=4):
$S_{ij}^{t_qt_j} = \frac{s_{ij}^{t_qt_g} + ... + s_{ij}^{t_q+K-1,t_g+K-1}}{K}$.
The sum in the numerator is the sum of diagonal elements started with $s_{ij}^{t_qt_j}$.

We define the intersection score between $(q_i, g_j)$ as 

$$
	\mathbf{S_{ij}} = \max_{\substack{a=1, ... ,s_i-K \\ b=1, ... , p_j-K}} S_{ij}^{ab}
$$

and the corresponding video segments as

\[
	(\mathbf{a}, \mathbf{a}+K), (\mathbf{b}, \mathbf{b}+K)
\]

where

\[
	\mathbf{a}, \mathbf{b} = \argmax_{\substack{a=1, ... ,s_i-K \\ b=1, ... , p_j-K}} S_{ij}^{ab}
\]

Finally we sorted all $\mathbf{S_{ij}}$ in the descending order and manually assess candidate pairs.

\subsubsection{Number of pairs to assess} \label{ssec:how_many_pairs}
Suppose we search duplicates in datasets Q and G and we have seen N pairs with the highest scores
and find M pairs with duplicates. The important question is: what is the total number of duplicates and how many
percents of them have we found.

For each pair of Q and G we construct the following test procedure. The first step is to augment Q, and let us call the result of augmentation as $\hat{\text{Q}}$.
To augment a dataset we apply two transformations: 1. we randomly crop a side of each video, where each side can be 70\%--100\% of original side length
(aspect ratio can be changed); 2. we randomly shift the start of the video by a random value between 0 and 1 seconds.

Having Q, $\hat{\text{Q}}$ and G we compute sets of positive and negative scores: Pos and Neg. The Pos is the set of scores between
i-th video from Q and the corresponding augmented video from $\hat{\text{Q}}$. Neg is the set of scores between
each video from Q and G. Having Pos and Neg sets we can plot a curve, where $x$ axis represents the fraction of found pairs with duplicates and Y axis
represents the number of negative pairs that we need to assess to find fraction $x$ of positive pairs, call this curve $F(x)$. 
We present the algorithm that computes $F$ using Pos and Neg sets in Lst.~\ref{lst:search_curve}.
Suppose we have seen $N+M$ pairs and have found $M$ pairs with duplicates. The total number of pairs with duplicates can be estimated as $M / F^{-1}(N)$.
By the definition $F(x)$ connects the fraction of found positive pairs with the number of seen negative pairs. The value $F^{-1}(N)$ represents approximation of the fraction of found
positive pairs. So if we know, that $M$ is approximately $100 * F^{-1}(N) \%$ of positive pairs, then we can approximately compute 100\% of positive pairs as $M / F^{-1}(N)$.

\begin{minipage}{\columnwidth}
\begin{lstlisting}[label=lst:search_curve, language=Python, caption={Numpy pseudocode for building the search curve $F(x)$},captionpos=b]
# first element is highest
P = np.sort(P)[::-1] # Pos
N = np.sort(N)[::-1] # Neg
xs = []
ys = []
for x, p in enumerate(P):
	# how many negative scores
	# greater than p ?
	j = np.searchsorted(N, p)
	xs.append(x)
	ys.append(j)
\end{lstlisting}
\end{minipage}

\subsubsection{Best 2D feature extractor}
The key component of a duplicate search system is a feature extractor. A good feature extractor significantly reduces the number of pairs for manual assessment.
To compare different 2D feature extractors we use the following test procedure. The test consists of two datasets.
The first dataset is the train part from the MSRVTT full split. The second dataset is random 596k videos
from the HowTo100M dataset. From each video of the taken part of HowTo100M we take a random 30 seconds segment.
We apply random augmentation to MSRVTT, as described in Sec.~\ref{ssec:how_many_pairs}.
Define MSRVTT as $Q$, the augmented MSRVTT dataset as $\hat{Q}$ and the taken part of HowTo100M as G.
For each feature extractor we compute curve $F(x)$, as described in Sec.~\ref{ssec:how_many_pairs}.

The best expert has the lowest curve. For example, if we want to find 95\% of duplicates, we should see many of candidates, some of them are duplicates,
but majority of them are not. So, the value $F(0.95)$ is the approximation of how many not duplicates we need to see to find 95\% of duplicates.
Ideally $F(0.95) = 0$, where all seen candidates are duplicates. So, a lower value $F(0.95)$ requires to see less number of false candidates, that's why
the lower curve is better.

\begin{figure}[h]
	\centering
			\begin{tabular}{@{}c@{}}
			\input{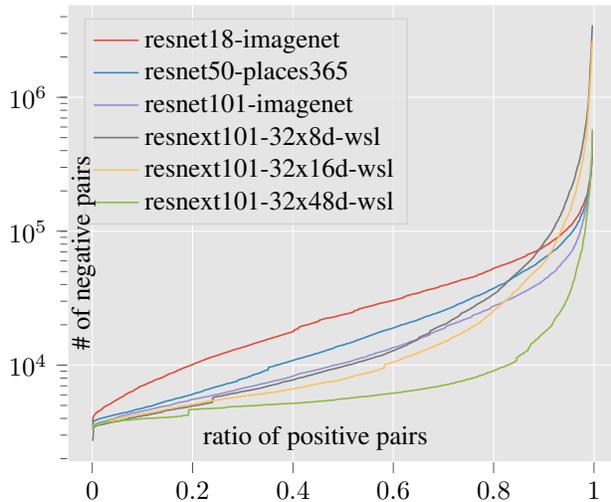} \lapbox[0pt]{-0.92\columnwidth}{\rotatebox{90}{\hspace{6em}\# of negative pairs}} \\
								   \raisebox{3.5em}[0pt][0pt]{ratio of positive pairs}
		\end{tabular}

	\caption{
		Search curves $F$ for different pretrained models.
		Curve F is used to estimate the minimal number of negative pairs (y = $F(x)$) that human assessors need to inspect before they find the fraction x of positive pairs. 
		The lower the curve $F$ the better (need to inspect manually less pairs).
		The curves are built with the query set Q = MSRVTT full train, the gallery set G = random 596k videos from HowTo100M.}
	\label{fig:search_curves_all}
\end{figure}

We consider several feature extractors: resnet18 and resnet101 \cite{he2015deep} pretrained on ImageNet \cite{imagenet_cvpr09},
resnet50 pretrained on Places365 \cite{zhou2017places} and resnext101-32x8d, resnext101-32x32d,
resnext101-32x48d pretrained on one billion images from Instagram \cite{wslimageseccv2018} and finetuned on ImageNet.
We report search curves $F(x)$ for these pretrained networks in Fig.~\ref{fig:search_curves_all}.

There exist networks \cite{radford2learning} \cite{kordopatis2017near} trained especially for match the
duplicate frames or video segments, but they are not publicly available.

As we see resnext101-32x48d-wsl shows the best result. We use this network for searching for duplicates.

It is worth to mention that here we just compare different networks on a fixed benchmark, and pick the best one. But the search curve $F(x)$ significantly depends on data.
This curve should be estimated for each used pair of datasets $Q$ and $G$.

\subsubsection{Black frames}
Often two consecutive video segments are glued with several black frames. The cosine similarity of embeddings of two black or near black frames are close to 1.
In this case the most probable candidates for duplicates are black video segments. To prevent this we apply the following rule. Suppose we have a frame $U$
and the unit length embedding $v$ computed from $U$. 
We find the prevalent color in $U$ and compute the area $S_0$ filled by this color. Then we compute
the value $S_0 / (hw)$, where $h$ and $w$ are the height and width of $U$. If this fraction is greater than 0.7 we define $\mu_v = 1 - S_0 / (hw)$,
otherwise $\mu_v = 1$. To calculate similarity between embeddings $v_1$ and $v_2$ we use weighted cosine similarity: $\mu_1\mu_2\text{cos}{v_1, v_2}$, instead of
classical cosine similarity. This rule removes majority of all near black frames from the most relevant candidates for duplicates.

\subsubsection{Screensavers detection}
Many videos from ActivityNet, HowTo100m, YouCook2 contain screensavers at the beginning or at the end. It causes a problem like mentioned above with near black frames,
because most of relevant proposals are the same screensavers, but the video content of the remainder video part are different.

Using the system described in Sec.~\ref{ssec:ndvs_gui} we search for duplicates in the ActivityNet dataset, where a lot of the most relevant segments are screensavers.
We collect several hundreds of screensavers and then compute embeddings for each of them. Let us call the resulting set of embeddings as E. Then we apply the following rule: if some embedding $v$ has the similarity
greater that 0.9 to one of embeddings from E, we set $v = 0$. So if the video segment has a part of a screensaver, it will never be in the most relevant proposals.

\subsubsection{GUI}\label{ssec:ndvs_gui}
\begin{figure}[h]
	\centering
	\includegraphics[scale=0.5]{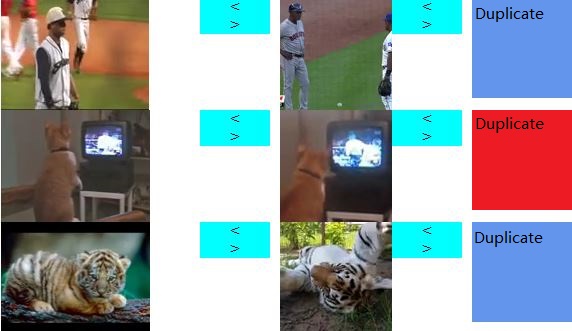}
	\caption{Web system used to find duplicates. Images on the first and third row are not duplicates and the second row contains duplicate.}
	\label{fig:ndvs_gui}
\end{figure}

The important part of the video duplicate search system is the user interface. Without ergonomic and fast interface it is impossible to
assess tens thousands of video pairs. Our system is presented in Fig.~\ref{fig:ndvs_gui}.

The system shows video pairs with the highest scores on top. A user needs to scroll down a web page (new videos are loaded dynamically with ajax), and if a video duplicate is detected,
a user should press the \textit{Duplicate} button, if there are no duplicates in the current viewport, no action is required. When a user scrolls a web page,
all non-duplicate pairs automatically are saved to a log file. Additionally several users at the same time can assess video pairs.

\begin{table}[H]
	\centering
	\begin{tabular}{
			|
			c@{\hspace{1\tabcolsep}}
			|
			c@{\hspace{1\tabcolsep}}
			c@{\hspace{1\tabcolsep}}
			|
			c@{\hspace{1\tabcolsep}}
			c@{\hspace{1\tabcolsep}}
			|
			c@{\hspace{1\tabcolsep}}
			c@{\hspace{1\tabcolsep}}
			|
		}
		\toprule
			\multirow{2}{*}{dataset} &
			\multicolumn{2}{c|}{M} & 
			\multicolumn{2}{c|}{A} & 
			\multicolumn{2}{c|}{L}
			\\
		& test & train & test & train & test & train \\
		\midrule
			M             & 114&223 &  6&10     & 0&0 \\
			A             & 10&6    &  127&163 & 0&0 \\
			L             & 6&2744   & 0&0      &	0&0 \\
			YouCook2      & 13&27    & 7&10     & 0&0 \\
			MSVD          & 1&1      & 1&1      & 1&1 \\
			TGIF          & 6&8      & 0&0      & 0&0 \\
			Twitter Vines & 3&3      & 0&0      & 0&0 \\
			Kinetics700   & 4&5      & 456&464 & 0&0 \\
			HowTo100M     & 177&154    & 209&209 & 0&0 \\
		\bottomrule
	\end{tabular}

	\caption{The leftmost column represents train parts of datasets, and the upper row represents test parts of datasets.
			 Column "test" means how many video segments are in the test part that have the corresponding pair in the train part either with the same YouTube ID or
			 manually marked as a duplicate. Column "train" represents the number of video segments in the train part that have
			 corresponding pair in the test dataset either with the same YouTube ID or manually marked as a duplicate. All segments counted in the "train" column
			 are removed from the train part. For example consider the column "A" and the row "M". train=10 means that the MSRVTT train part contains 10 video segments that
			 have a pair in the ActivityNet test part. These 10 videos must be removed from train part when dataset are combined.			 
			 test=6 means that ActivityNet test has 6 video segments that have a pair in the MSRVTT test part.
			 }
	\label{tab:found_isect}
\end{table}
\subsection{Cleaning results}\label{sect:cleaning_res}

Recall that our cleaning method consists of two stages. In the first stage we throw out from the train part all video segments that have a pair with the same YouTube ID
in test parts of MSRVTT or ActivityNet. The second stage is matching video segments by embeddings and manually assess several thousands
pairs with the highest score.

In Tab.~\ref{tab:found_isect} we report how many duplicates are found for each pairs of datasets. This table represents the final
result after applying these two stages.

Separate results for the first and the second stages are reported in Sec.~\ref{sect:isect_ytvid}.

Note that columns "test" and "train" in Tab.~\ref{tab:found_isect} may have different values. Consider the situation when the test part have a video segment A, and
the train part have two video segments A1 and A2. And both are marked as duplicates with A. In this case the video segment A brings +1 to the "test" column
and A1, A2 bring +2 to the "train" column.

The most problematic datasets in terms of the number of duplicates are
MSRVTT and ActivityNet. These datasets overlap with itself (e.g. MSRVTT test overlap with MSRVTT train).
We found more than 100 duplicate pairs for both of them.
Other problematic datasets are HowTo100M and Kinetics700, these datasets are large, so we can't assess the required number of video pairs to find 95\% or 99\%
of duplicates. But we can assess a smaller number of pairs and using search curves $F$ (see Sec.~\ref{ssec:how_many_pairs}) can extrapolate this value to 100\%.
We found that HowTo100M may have the intersection with MSRVTT test full by about 300 videos (10\% of the MSRVTT test full).
The similar situation is about the ActivityNet test set and Kinetics700, the intersection could be near 500-600 videos (10\% of the ActivityNet test set).

In Tab.~\ref{tab:mmt_cleaning} we report results on MSRVTT for MMT retraining with no cleaning, after cleaning by the YouTube ID and cleaning combination by the YouTube ID and the manual assessment.
The manual cleaning for 1k-A and 1k-B is incomplete because we only do cleaning for the full split. The following situation takes place for 1k-A, 1k-B splits: when 1k videos from the full test are taken for test
and the remaining 2k videos are moved to the train part, the additional overlapping is introduced, because these 1k and 2k videos are overlapping. We do not remove this overlap in this research.

\begin{table}[H]
	\centering
	\begin{tabular}{|ll@{\hspace{1\tabcolsep}}l@{\hspace{1\tabcolsep}}l|}
		\toprule
		\multirow{2}{*}{split} & no    & \multirow{2}{*}{by ID} & by ID + \\
							   & clean &                        & manual \\

		\midrule
		full &  31.1$_{\pm 0.1}$ &  31.1$_{\pm 0.1}$ &  30.2$_{\pm 0.4}$\\
		1k-A &  54.8$_{\pm 0.5}$ &  50.7$_{\pm 0.9}$ &  49.4$_{\pm 0.5}$\\
		1k-B &  51.1$_{\pm 0.9}$ &  46.1$_{\pm 0.1}$ &  46.4$_{\pm 0.6}$\\
		\bottomrule
	\end{tabular}

	\caption{Comparison for original MMT trained (7 modalities) on MSRVTT without cleaning, with cleaning by the YouTube ID only, and
			 with cleaning by the YouTube ID plus the manual assessment.}
	\label{tab:mmt_cleaning}
\end{table}

As you can see after cleaning the performance is significantly decreased on 1k-A and 1k-B splits for original MMT.

\subsubsection{Intersection by YouTube ID and embeddings}\label{sect:isect_ytvid}

In Tab.~\ref{tab:ytvid_isect} we report the intersection by the YouTube ID between test parts of MSRVTT (full, 1k-A, 1k-B) and ActivityNet with
train parts of MSRVTT (full, 1k-A, 1k-B), ActivityNet, Kinetics700, YouCook2, HowTo100m, MSVD.

\begin{table}[ht]
	\centering
	\begin{tabular}{
			|@{\hspace{0.5\tabcolsep}}
			c@{\hspace{0.5\tabcolsep}}
			|@{\hspace{0.5\tabcolsep}}
			c@{\hspace{1\tabcolsep}}
			c@{\hspace{0.5\tabcolsep}}
			|@{\hspace{0.5\tabcolsep}}
			c@{\hspace{1\tabcolsep}}
			c@{\hspace{0.5\tabcolsep}}
			|@{\hspace{0.5\tabcolsep}}
			c@{\hspace{1\tabcolsep}}
			c@{\hspace{0.5\tabcolsep}}
			|@{\hspace{0.5\tabcolsep}}
			c@{\hspace{1\tabcolsep}}
			c@{\hspace{0.5\tabcolsep}}
			|
		}
		\toprule
		data &
		\multicolumn{2}{c|@{\hspace{0.5\tabcolsep}}}{M} &
		\multicolumn{2}{c|@{\hspace{0.5\tabcolsep}}}{M$_{\text{1k-a}}$} &
		\multicolumn{2}{c|@{\hspace{0.5\tabcolsep}}}{M$_{\text{1k-b}}$} &
		\multicolumn{2}{c|}{A}
		\\
		set &
		{\footnotesize test} & {\footnotesize train} & 
		{\footnotesize test} & {\footnotesize train} & 
		{\footnotesize test} & {\footnotesize train} &
		{\footnotesize test} & {\footnotesize train}
		\\
		\midrule
		M          	       & 0&0 & 0&0  & 104&179 & 2&4     \\
		M$_{\text{1k-a}}$  & 2362&1990  & 372&415 & 827&1007 & 2&4     \\ 
		M$_{\text{1k-b}}$  & 1689&1367  & 563&634 & 380&407  & 2&4     \\
		A	    	       & 0&0        & 0&0     & 0&0      & 0&0     \\
		K        		   & 5&4        & 1&1     & 0&0      & 408&408 \\
		Y       		   & 8&4        & 2&2     & 2&2      & 3&3     \\
		HT100M   		   & 147&117    & 39&38   & 57&53    & 175&175 \\
		MSVD		       & 3&1        & 2&1     & 0&0      & 1&1     \\
		\bottomrule
	\end{tabular}

	\caption{First stage. The leftmost column represents train parts of datasets, and the upper row represents test parts of datasets. Column "test" represents
	number of video segments in test part that have corresponding video in train part with the same YouTube ID. Column "train" represents number of video
	in train part that have corresponding pair in test part with the same ID. For example: if we combine M and YouCook2, we should remove 4 video from YouCook2 train.}
	\label{tab:ytvid_isect}
\end{table}

It is worth to mention that MSRVTT 1k-A test and 1k-B test have a large overlap ratio by the YouTube ID with the 1k-A train and the 1k-B train parts correspondingly. 
Both splits have the overlap ratio of about 38\% between the train part and the test part.
We also emphasize that the original MSRVTT full split does not overlap by the YouTube ID between the test and train parts.

\begin{table}[ht]
	\centering
	\begin{tabular}{
			|
			c
			|@{\hspace{0.5\tabcolsep}}
			c@{\hspace{0.5\tabcolsep}}
			c@{\hspace{0.5\tabcolsep}}
			c@{\hspace{0.5\tabcolsep}}
			|@{\hspace{0.5\tabcolsep}}
			c@{\hspace{0.5\tabcolsep}}
			c@{\hspace{0.5\tabcolsep}}
			c@{\hspace{0.5\tabcolsep}}
			|@{\hspace{0.5\tabcolsep}}
			c@{\hspace{0.5\tabcolsep}}
			c@{\hspace{0.5\tabcolsep}}
			c@{\hspace{0.5\tabcolsep}}
			|
		}
		\toprule
			data &
			\multicolumn{3}{c| @{\hspace{0.5\tabcolsep}}}{M} & 
			\multicolumn{3}{c| @{\hspace{0.5\tabcolsep}}}{A} & 
			\multicolumn{3}{c|}{L}
			\\
		set &  
		{\footnotesize seen} &
		{\footnotesize found} &
		{\footnotesize total} & 
		{\footnotesize seen} &
		{\footnotesize found} &
		{\footnotesize total} & 
		{\footnotesize seen} &
		{\footnotesize found} &
		{\footnotesize total}
		\\
		\midrule
			M             & 10k&114&114 &    1k&      6&6 &   1k&0&0 \\ 
			A             & 10k&10&10   &    15k& 127&142 &   1k&0&0 \\
			L             & 3k& 6&6     &    2k&      0&0 &   ---&---&--- \\
			Y             & 2k& 13&13   &    1k&      7&7 &   1k&0&0 \\
			MSVD          & 1k& 1&1     &    1k&      1&1 &   1k&1&1 \\
			T             & 2k& 6&6     &    2k&      0&0 &   3k&0&0 \\
			V             & 2k& 3&3     &    0k&      0&0 &   1k&0&0 \\
			K             & 2k& 1&2     &    30k& 227&539 &   2k&0&0 \\
			HT100M        & 5k& 15&320  &    ---& ---&--- &   ---&---&--- \\
		\bottomrule
	\end{tabular}
	
	\caption{Second stage. The leftmost column represents train parts of datasets, and the upper row represents test parts of datasets.
			 Column "seen" represents the number of video segments that we manually assess for a given pair of datasets.
	         Column "found" represents the number of videos in the test part for which there exists the corresponding duplicate video segment in the train part.
			 Column "total" represents the approximately estimated total number of videos from the test part that have a duplicate pair in the train part.
			 Symbol "---" means that the intersection is not computed because it requires too much human resources.}
	\label{tab:manual_isect_Ntest}
\end{table}

In Tab.~\ref{tab:manual_isect_Ntest} we report the statistics for the second deduplication stage (searching by embeddings).
We do not compute an intersection for MSRVTT 1k-A and 1k-B splits. 

In this table we present the number of manually found duplicates and the estimated maximum number of duplicates for a given pair of datasets.
We managed to find the intersection for almost all pairs of datasets.

The maximum number of duplicates is computed based on the search curve $F(x)$. As we told in Sec.~\ref{ssec:how_many_pairs} the search curve significantly depends on data. We compute
the search curve for all pairs of datasets in Tab.~\ref{tab:manual_isect_Ntest}. The search curve for each particular pair of datasets is build exactly in the same way
as described in Sec.~\ref{ssec:how_many_pairs}. For example, to compute the search curve for MSRVTT test and ActivityNet train we define MSRVTT test as $Q$, ActivityNet train as $G$, then augment
$Q$ to produce $\hat{Q}$, and use the algorithm described in Sec.~\ref{ssec:how_many_pairs}.

Using the column "seen" from Tab.~\ref{tab:manual_isect_Ntest} we can compute how many pairs need to be assessed to find the full overlap between datasets. For example, inspect
5k pairs for HowTo100M dataset and MSRVTT (the row "HT100M" and the column "M"), we found 15 duplicates, so the approximate maximum number of duplicates is 320: 5k * (320 / 15) = 106k.
So, to find the full overlap using the current version of algorithm it is needed to manually assess 106k video pairs and it is too much, that's why we do not find full intersection
for this specific pair of datasets.

\section{Hyperparameters}
To train our best networks (MMT(MALVYMTS) L9H8 CLIP+audio,  MDMMT(MALVYMTS) L9H8 irCSN152+audio and MMT(MALVYMTS) L9H8 CLIP+irCSN152+audio)
we use 50 epochs and define a single epoch as 150K examples per
GPU (in total 1.2M examples per epoch on 8 GPUs). We use Adam optimizer without weight decay, the initial value for a learning rate is 5e-5, after each epoch we multiply the learning rate by 0.95.
Batch size of 32 examples per GPU is used. We do not exchange embeddings between GPUs.
We use bi-directional max-margin ranking loss with margin 0.05.
In Bert and the video transformer encoder we use dropout 0.2 in attention and in FFN block. We use 8 Nvidia V100 32GB GPUs. The training time is about 14 hours.

\section{Pretrained model}\label{sect:pretrain_ht100m}
The well known method to boost the performance in video retrieval tasks is to use a pretrained model. First the neural network is trained on some large dataset, then
at second stage it is finetuned for target target dataset. In video retrieval task HowTo100M dataset is often used for pretraining. In this work we use HowTo100M for pretraining in the same way.

\begin{table*}[htb!]
  \centering
  \begin{tabular}{|l @{\hspace{1\tabcolsep}} |l@{\hspace{1\tabcolsep}}| l@{\hspace{1\tabcolsep}}l@{\hspace{1\tabcolsep}}l@{\hspace{1\tabcolsep}}l@{\hspace{1\tabcolsep}}l|}
    \toprule
    \multirow{2}{*}{model} &\multirow{2}{*}{\rotatebox{90}{pretr}}   & \multicolumn{5}{c|}{MSRVTT full clean  text $\rightarrow$ video} \\
			   && R@1$\uparrow$ & R@5$\uparrow$ & R@10$\uparrow$ & MnR$\downarrow$ & MdR$\downarrow$ \\
    \midrule
      \ours MDMMT(M$_c$ALVYMTS) L9H8 irCSN152+audio	    &yes& 15.8$_{\pm 0.1}$ & 38.9$_{\pm 0.1}$ & 51.0$_{\pm 0.1}$ & 76.4$_{\pm 0.5}$ & 10.0$_{\pm 0.0}$ \\
      \ours MDMMT(M$_c$ALVYMTS) L9H8 irCSN152+audio	    &no & 14.5$_{\pm 0.1}$ & 36.8$_{\pm 0.3}$ & 48.8$_{\pm 0.3}$ & 82.2$_{\pm 0.6}$ & 11.0$_{\pm 0.0}$\\
      \ours MDMMT(M$_c$ALVYMTS) L9H8 CLIP+audio             &yes& 21.5$_{\pm 0.1}$ & 47.4$_{\pm 0.2}$ & 59.6$_{\pm 0.1}$ & 57.7$_{\pm 0.4}$ & 6.0$_{\pm 0.0}$ \\
      \ours MDMMT(M$_c$ALVYMTS) L9H8 CLIP+audio             &no & 20.0$_{\pm 0.1}$ & 45.1$_{\pm 0.1}$ & 57.3$_{\pm 0.1}$ & 63.1$_{\pm 0.1}$ & 7.0$_{\pm 0.0}$ \\
    \bottomrule
  \end{tabular}
  \caption{Performance on the MSRVTT full clean split with and without pretrained model (HowTo100m).}
  \label{tab:pretr-msrvtt}

\vspace{\textfloatsep}

  \begin{tabular}{|l @{\hspace{1\tabcolsep}} |l@{\hspace{1\tabcolsep}}| l@{\hspace{1\tabcolsep}}l@{\hspace{1\tabcolsep}}l@{\hspace{1\tabcolsep}}l@{\hspace{1\tabcolsep}}l|}
    \toprule
    \multirow{2}{*}{model} &\multirow{2}{*}{\rotatebox{90}{pretr}}   & \multicolumn{5}{c|}{ActivityNet  text $\rightarrow$ video} \\
			   && R@1$\uparrow$ & R@5$\uparrow$ & R@10$\uparrow$ & MnR$\downarrow$ & MdR$\downarrow$ \\
    \midrule
      \ours MDMMT(M$_c$ALVYMTS) L9H8 irCSN152+audio &yes& 15.1$_{\pm 0.1}$ & 38.3$_{\pm 0.1}$ & 51.5$_{\pm 0.3}$ & 92.4$_{\pm 2.3}$ & 10.0$_{\pm 0.0}$ \\
      \ours MDMMT(M$_c$ALVYMTS) L9H8 irCSN152+audio &no & 12.0$_{\pm 0.1}$ & 33.7$_{\pm 0.4}$ & 46.3$_{\pm 0.3}$ & 119.9$_{\pm 2.1}$ & 13.0$_{\pm 0.0}$ \\
      \ours MDMMT(M$_c$ALVYMTS) L9H8 CLIP+audio     &yes& 17.7$_{\pm 0.1}$ & 41.6$_{\pm 0.3}$ & 54.3$_{\pm 0.2}$ & 76.0$_{\pm 1.0}$ & 8.3$_{\pm 0.5}$ \\
      \ours MDMMT(M$_c$ALVYMTS) L9H8 CLIP+audio     &no & 15.2$_{\pm 0.3}$ & 37.9$_{\pm 0.3}$ & 50.1$_{\pm 0.2}$ & 93.4$_{\pm 2.0}$ & 10.3$_{\pm 0.5}$ \\
    \bottomrule
  \end{tabular}
  \caption{Performance on ActivityNet with and without pretrained model (HowTo100m). The performance reported for the text to video retrieval task on our own subset of the original ActivityNet test part.
	   See Sec.~\ref{sect:datasets} for details.}
  \label{tab:pretr-anet}

\vspace{\textfloatsep}

  \begin{tabular}{|l @{\hspace{1\tabcolsep}} |l@{\hspace{1\tabcolsep}}| l@{\hspace{1\tabcolsep}}l@{\hspace{1\tabcolsep}}l@{\hspace{1\tabcolsep}}l@{\hspace{1\tabcolsep}}l|}
    \toprule
    \multirow{2}{*}{model} &\multirow{2}{*}{\rotatebox{90}{pretr}}   & \multicolumn{5}{c|}{LSMDC  text $\rightarrow$ video} \\
			   && R@1$\uparrow$ & R@5$\uparrow$ & R@10$\uparrow$ & MnR$\downarrow$ & MdR$\downarrow$ \\
    \midrule
      \ours MDMMT(M$_c$ALVYMTS) L9H8 irCSN152+audio &yes& 13.1$_{\pm 0.5}$ & 31.3$_{\pm 0.3}$ & 40.1$_{\pm 0.0}$ & 74.5$_{\pm 0.7}$ & 19.3$_{\pm 0.5}$\\
      \ours MDMMT(M$_c$ALVYMTS) L9H8 irCSN152+audio &no&  12.6$_{\pm 0.7}$ & 30.2$_{\pm 1.5}$ & 39.6$_{\pm 0.9}$ & 76.1$_{\pm 0.8}$ & 19.7$_{\pm 1.3}$\\
      \ours MDMMT(M$_c$ALVYMTS) L9H8 CLIP+audio     &yes& 17.2$_{\pm 0.6}$ & 34.9$_{\pm 0.4}$ & 45.3$_{\pm 1.0}$ & 65.6$_{\pm 0.8}$ & 14.0$_{\pm 0.8}$\\
      \ours MDMMT(M$_c$ALVYMTS) L9H8 CLIP+audio     &no&  16.2$_{\pm 1.1}$ & 35.4$_{\pm 1.3}$ & 45.1$_{\pm 0.7}$ & 64.9$_{\pm 1.9}$ & 14.7$_{\pm 0.5}$\\
    \bottomrule
  \end{tabular}
  \caption{Performance on LSMDC with and without pretrained model (HowTo100m).}
  \label{tab:pretr-lsmdc}
\end{table*}

In our training procedure we use 8 Nvidia V100 32Gb GPUs, we train for 200 epochs where one epoch is defined as 80k examples on each GPU (in total network sees 640k examples on 8 GPUs per epoch).
We use batch size 64 for each GPU and do not exchange embeddings between GPU.
Initial learning rate is 5e-5. After each epoch we multiply learning rate by 0.98.
We use the full HowTo00M dataset. The model is trained either with two modalities: motion/RGB and audio
or with three modalities: motion, RGB and audio, depending on how many modalities are used in final model.
The total training time is about 24 hours. We use bi-directional max-margin ranking loss with margin 0.05.

In Tab.~\ref{tab:pretr-msrvtt},~\ref{tab:pretr-anet}~and~\ref{tab:pretr-lsmdc} we compare two our models: MDMMT(M$_c$ALVYMTS) L9H8 irCSN152+audio and MDMMT(M$_c$ALVYMTS) L9H8 CLIP+audio
when they are trained from the pretrained model or not. In these three tables we present the same four models (no special finetuning for the target dataset) tested on different datasets.

As we can see in Tab.~\ref{tab:pretr-msrvtt} the pretrained model increases R1 metric by 1\% and R5 by 2\%. The pretrained model also increase performance on ActivityNet dataset, see Tab.~\ref{tab:pretr-anet}.
For R1 metric the improvement is about 2\% and for R5 metric is about 4\%. For LSMDC dataset, see Tab~\ref{tab:pretr-lsmdc}, we have approximately the same results with and without pretraining.


\end{document}